\documentclass[10pt,twocolumn,letterpaper]{article}

\usepackage{3dv}
\usepackage{times}
\usepackage{epsfig}
\usepackage{graphicx}
\usepackage{amsmath}
\usepackage{amssymb}
\usepackage{caption}

\usepackage{acronym}
\usepackage{comment}
\usepackage{booktabs}
\usepackage{float}
\usepackage{rotating}
\usepackage{makecell}

\usepackage[pagebackref=true,breaklinks=true,letterpaper=true,colorlinks,bookmarks=false]{hyperref}

\threedvfinalcopy 

\def\threedvPaperID{164} 

\ifthreedvfinal\fi

\newcommand{\qheading}[1]{\noindent\textbf{#1}}

\newcommand{\ourmodel}{GIF }

\acrodef{PCA}{Principal Component Analysis}
\acrodef{GAN}{generative adversarial network}
\acrodef{AMT}{Amazon Mechanical Turk}

\begin{document}

\title{GIF: Generative Interpretable Faces}

\author{Partha Ghosh$^1$ \hspace{.4 cm} Pravir Singh Gupta*$^2$ \hspace{.4 cm} Roy Uziel*$^3$ \\ Anurag Ranjan$^1$ \hspace{.4 cm} Michael J. Black$^1$ \hspace{.4 cm} Timo Bolkart$^1$\\[5pt]
$^1$\ Max Planck Institute for Intelligent Systems, T\"ubingen, Germany\\
\textrm{$^2$\ Texas A\&M University, College Station, USA} \qquad \textrm{$^3$\ Ben Gurion University, Be'er Sheva, Israel}\\
{\tt\small \{pghosh,aranjan,black,tbolkart\}@tue.mpg.de \quad pravir@tamu.edu \quad uzielr@post.bgu.ac.il}
}

\twocolumn[{%
	\renewcommand\twocolumn[1][]{#1}%
	\maketitle
	\begin{center}
        \centerline{
            \includegraphics[height=0.058\textheight]{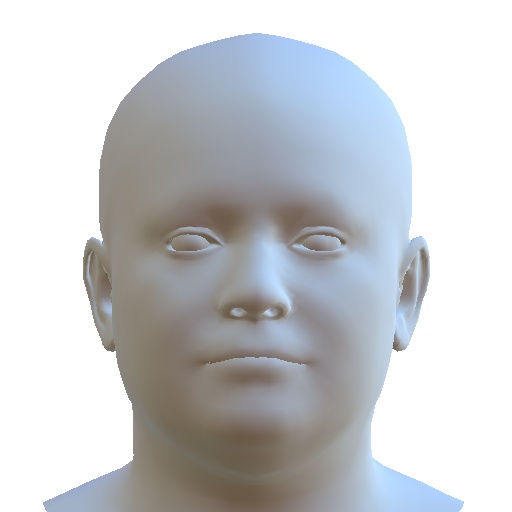}
            \includegraphics[height=0.058\textheight]{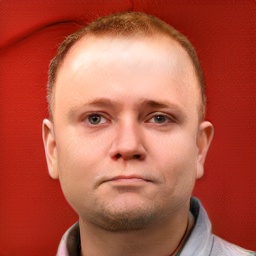}
            \includegraphics[height=0.058\textheight]{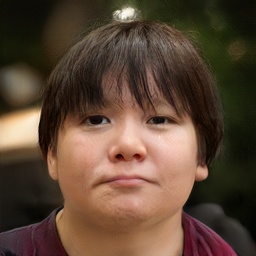}
            \hspace{0.01\textwidth}
            \includegraphics[height=0.058\textheight]{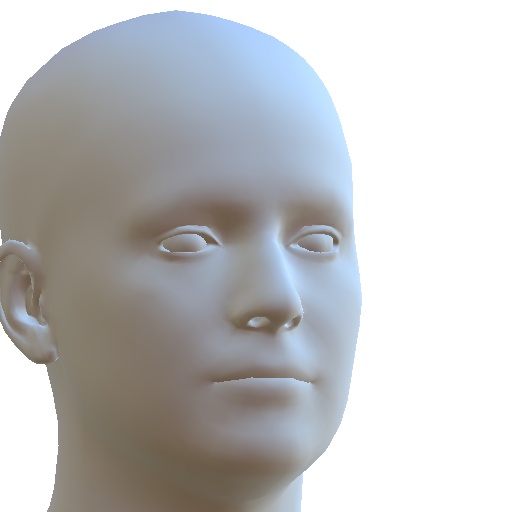}
            \includegraphics[height=0.058\textheight]{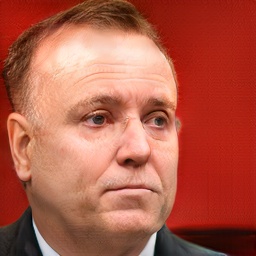}
            \includegraphics[height=0.058\textheight]{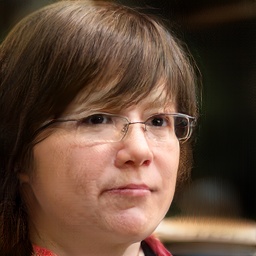}
            \hspace{0.01\textwidth}
            \includegraphics[height=0.058\textheight]{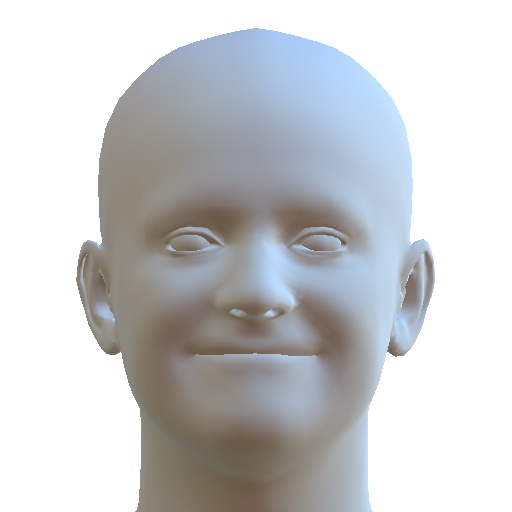}
            \includegraphics[height=0.058\textheight]{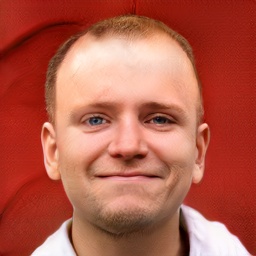}
            \includegraphics[height=0.058\textheight]{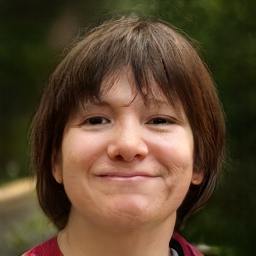}
            \hspace{0.01\textwidth}
            \includegraphics[height=0.058\textheight]{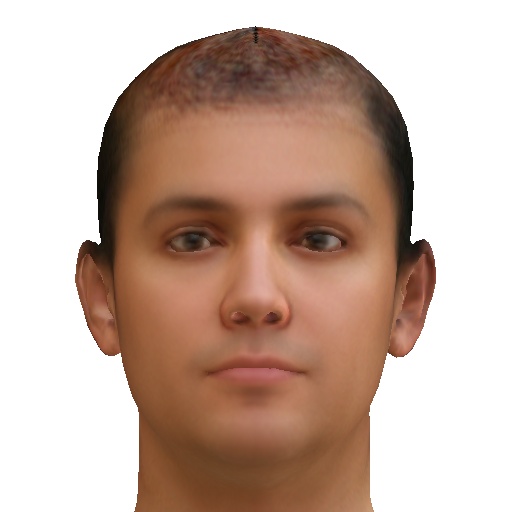}
            \includegraphics[height=0.058\textheight]{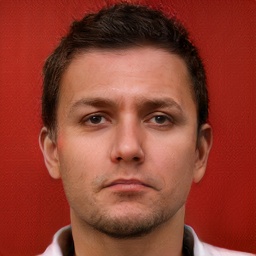}
            \includegraphics[height=0.058\textheight]{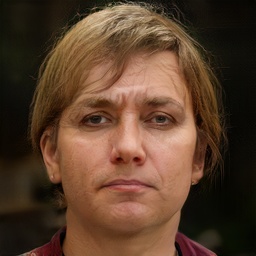}
        }
        \centerline{
            \includegraphics[height=0.058\textheight]{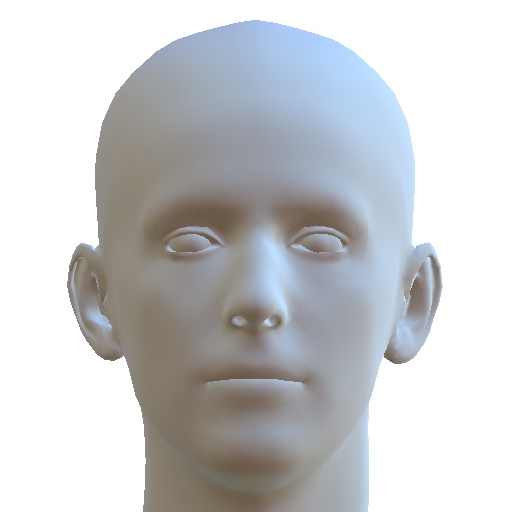}
            \includegraphics[height=0.058\textheight]{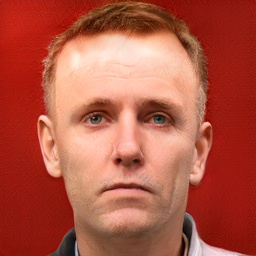}
            \includegraphics[height=0.058\textheight]{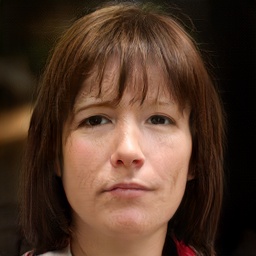}        
            \hspace{0.01\textwidth}
            \includegraphics[height=0.058\textheight]{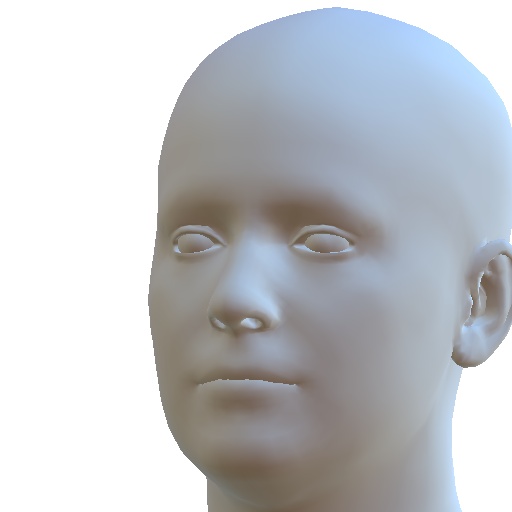}
            \includegraphics[height=0.058\textheight]{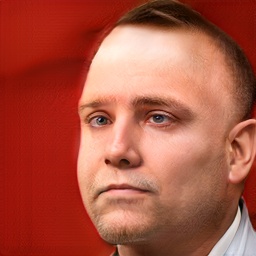}
            \includegraphics[height=0.058\textheight]{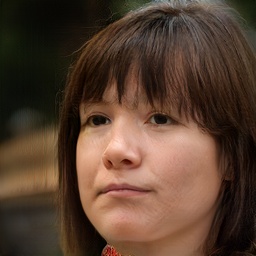}
            \hspace{0.01\textwidth}
            \includegraphics[height=0.058\textheight]{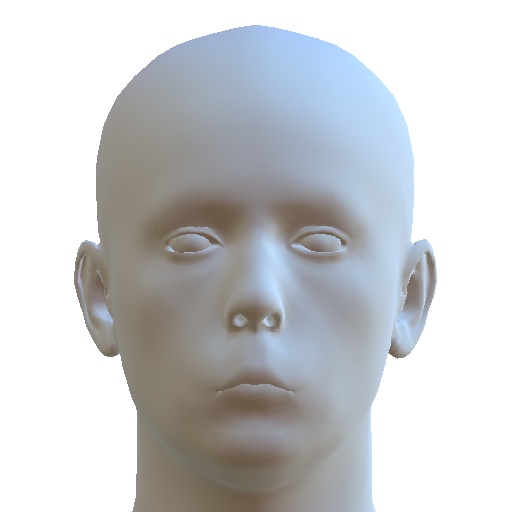}
            \includegraphics[height=0.058\textheight]{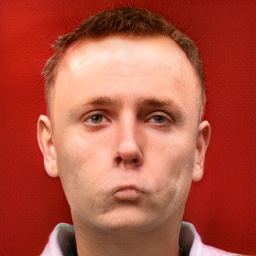}
            \includegraphics[height=0.058\textheight]{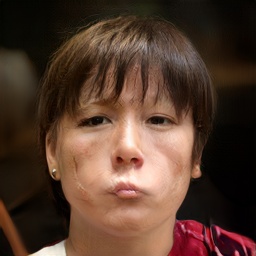}
            \hspace{0.01\textwidth}
            \includegraphics[height=0.058\textheight]{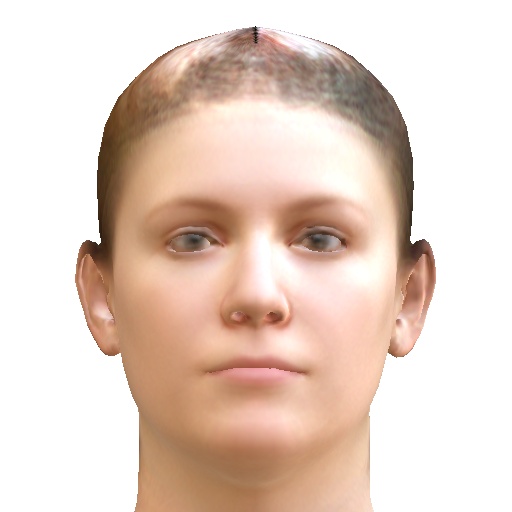}
            \includegraphics[height=0.058\textheight]{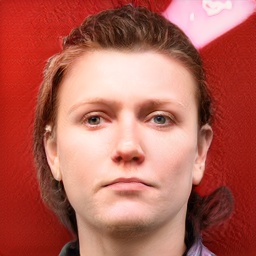}
            \includegraphics[height=0.058\textheight]{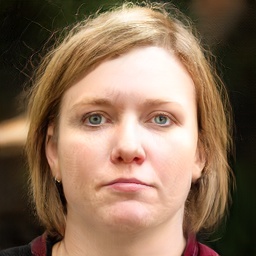} 
        }
        \centerline{
            \includegraphics[height=0.058\textheight]{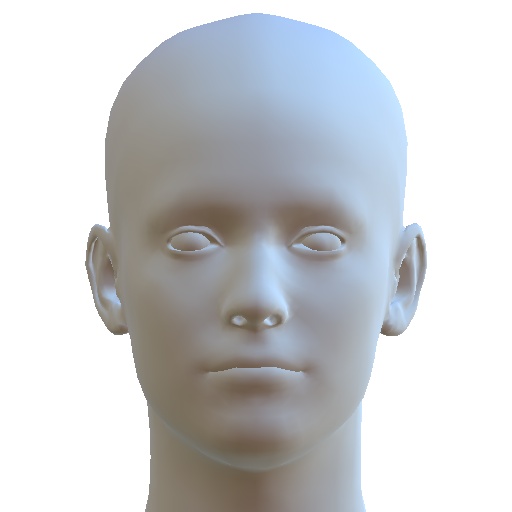}
            \includegraphics[height=0.058\textheight]{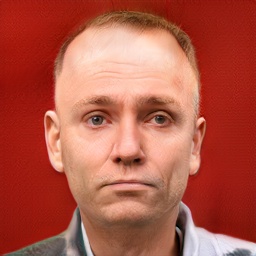}
            \includegraphics[height=0.058\textheight]{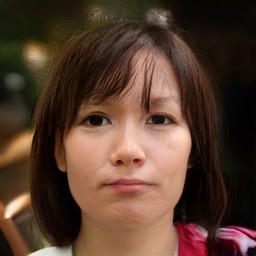}        
            \hspace{0.01\textwidth}
            \includegraphics[height=0.058\textheight]{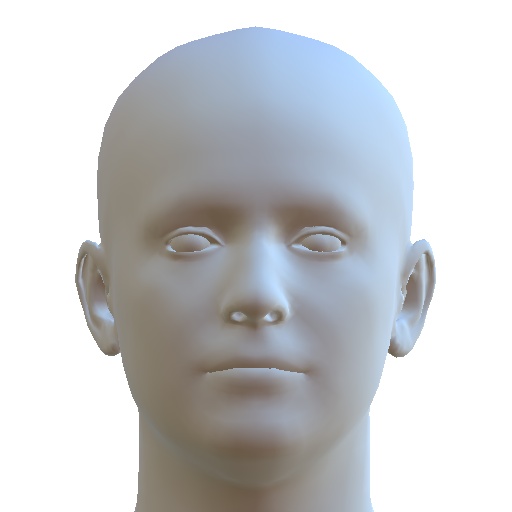}
            \includegraphics[height=0.058\textheight]{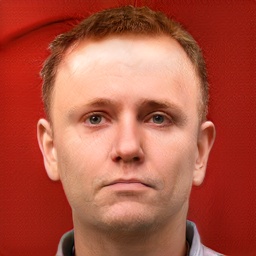}
            \includegraphics[height=0.058\textheight]{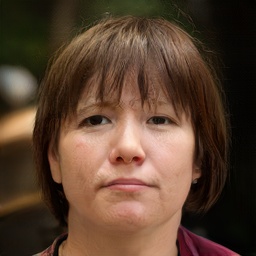}
            \hspace{0.01\textwidth}
            \includegraphics[height=0.058\textheight]{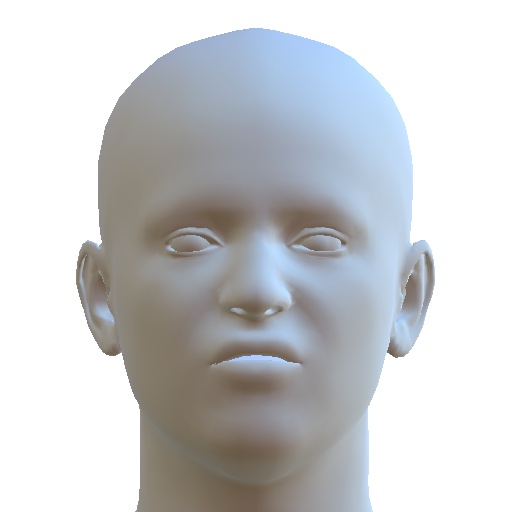}
            \includegraphics[height=0.058\textheight]{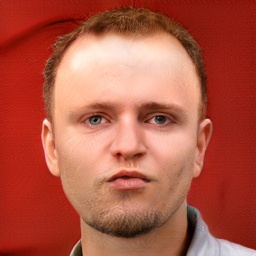}
            \includegraphics[height=0.058\textheight]{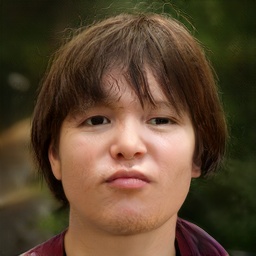}
            \hspace{0.01\textwidth}
            \includegraphics[height=0.058\textheight]{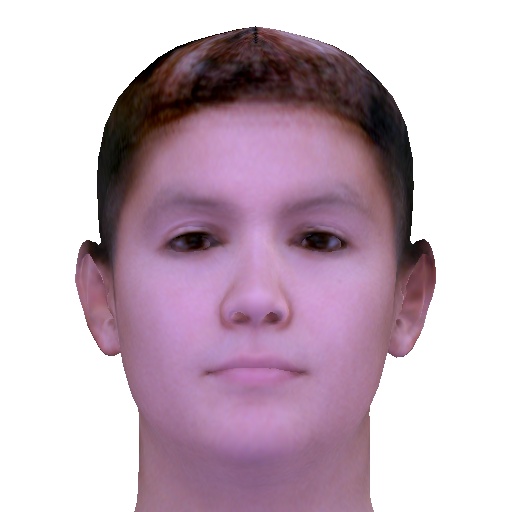}
            \includegraphics[height=0.058\textheight]{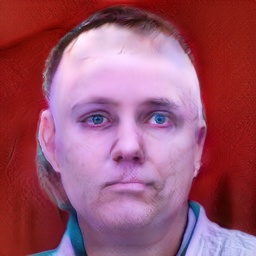}
            \includegraphics[height=0.058\textheight]{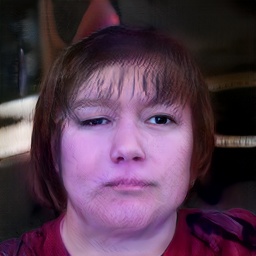}
        }
        \centerline{
            \includegraphics[height=0.058\textheight]{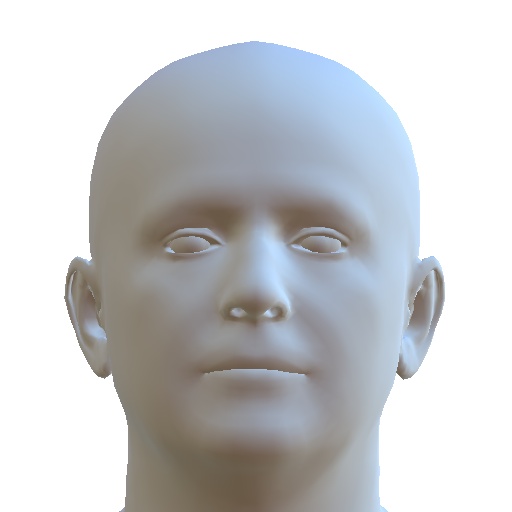}
            \includegraphics[height=0.058\textheight]{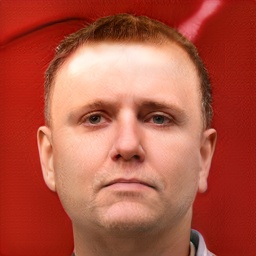}
            \includegraphics[height=0.058\textheight]{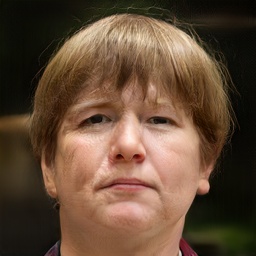}
            \hspace{0.01\textwidth}
            \includegraphics[height=0.058\textheight]{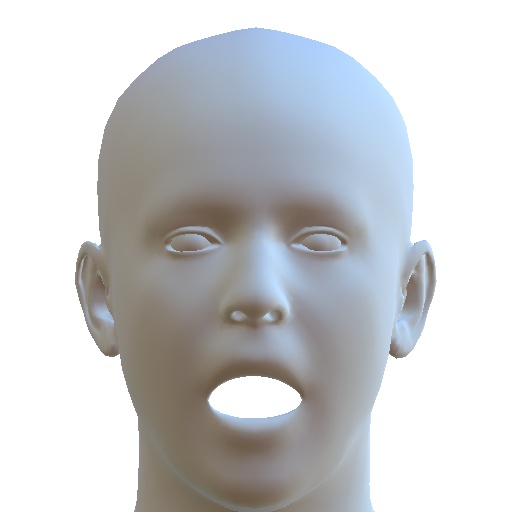}
            \includegraphics[height=0.058\textheight]{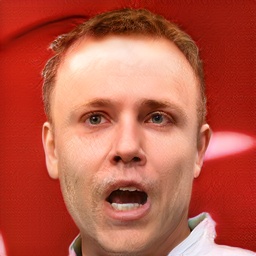}
            \includegraphics[height=0.058\textheight]{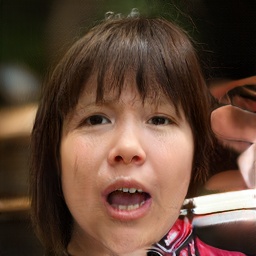}
            \hspace{0.01\textwidth}
            \includegraphics[height=0.058\textheight]{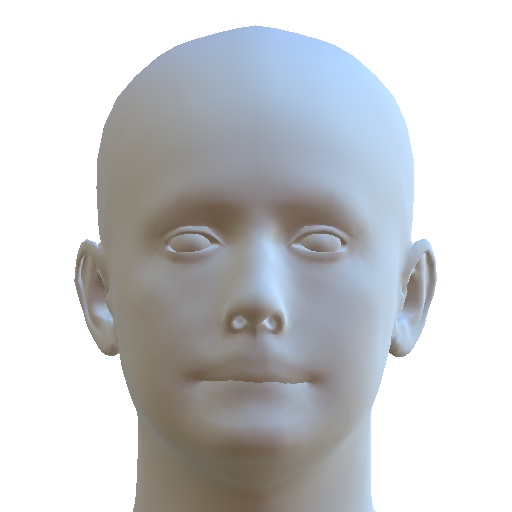}
            \includegraphics[height=0.058\textheight]{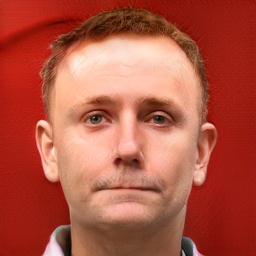}
            \includegraphics[height=0.058\textheight]{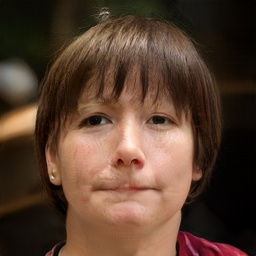}
            \hspace{0.01\textwidth}
            \includegraphics[height=0.058\textheight]{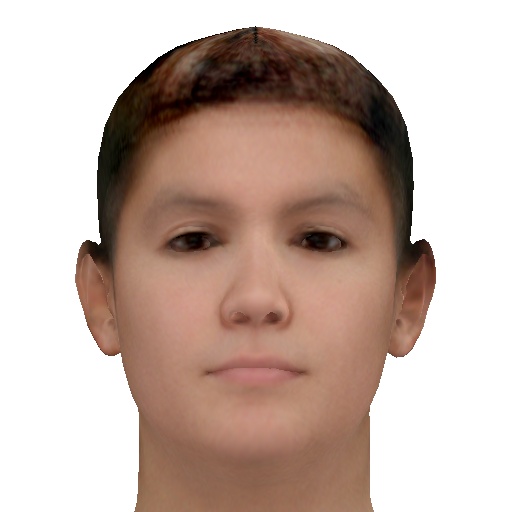}
            \includegraphics[height=0.058\textheight]{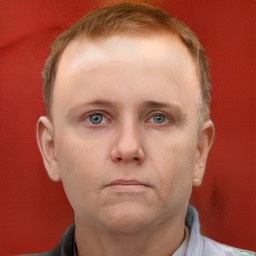}
            \includegraphics[height=0.058\textheight]{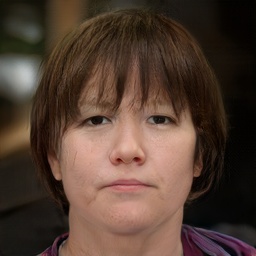}
        }
        \hspace{0.09\textwidth} Shape \hspace{0.2\textwidth} Pose \hspace{0.18\textwidth} Expression \hspace{0.11\textwidth} Appearance / Lighting \hspace{\fill}
    \captionof{figure}{Face images generated by controlling FLAME~\cite{FLAME:SiggraphAsia2017} parameters, appearance parameters, and lighting parameters. 
    For shape and expression, two principal components are visualized at $\pm 3$ standard deviations. The pose variations are visualized at $\pm \pi/8$ (head pose) and at $0, \pi/12$ (jaw pose). 
    For shape, pose, and expression, the two columns are generated for two randomly chosen sets of appearance, lighting, and style parameters.
    For the appearance and lighting variations (right), the top two rows visualize the first principal components of the appearance space at $\pm 3$ standard deviations, the bottom two rows visualize the first principal component of the lighting parameters at $\pm 2$ standard deviations.
    The two columns are generated for two randomly chosen style parameters.
    }
    \label{fig:teaser_with_lighting}
    \end{center}
}]

\newcommand{\gif}{GIF}
\newcommand{\shapecoeff}{\boldsymbol{\beta}}
\newcommand{\shapedim}{{\left| \shapecoeff \right|}}

\newcommand{\posecoeff}{\boldsymbol{\theta}}
\newcommand{\posedim}{{\left| \posecoeff \right|}}

\newcommand{\expcoeff}{\boldsymbol{\psi}}
\newcommand{\expdim}{{\left| \expcoeff \right|}}

\newcommand{\albedocoeffs}{\boldsymbol{\alpha}}
\newcommand{\albedodim}{{\left| \albedocoeffs \right|}}

\newcommand{\lighting}{\mathbf{l}}
\newcommand{\style}{\mathbf{s}}
\newcommand{\camera}{\mathbf{c}}

\newcommand{\numverts}{N}
\newcommand{\imgsize}{P}

\newcommand{\discriminator}{{D}}
\newcommand{\generator}{{G}}

\newcommand{\expt}{\mathbb{E}}
\newcommand{\Expt}{\expt}
\newcommand{\EXPT}{\expt}

\def\reta{{\textnormal{$\eta$}}}
\def\ra{{\textnormal{a}}}
\def\rb{{\textnormal{b}}}
\def\rc{{\textnormal{c}}}
\def\rd{{\textnormal{d}}}
\def\re{{\textnormal{e}}}
\def\rf{{\textnormal{f}}}
\def\rg{{\textnormal{g}}}
\def\rh{{\textnormal{h}}}
\def\ri{{\textnormal{i}}}
\def\rj{{\textnormal{j}}}
\def\rk{{\textnormal{k}}}
\def\rl{{\textnormal{l}}}
\def\rn{{\textnormal{n}}}
\def\ro{{\textnormal{o}}}
\def\rp{{\textnormal{p}}}
\def\rq{{\textnormal{q}}}
\def\rr{{\textnormal{r}}}
\def\rs{{\textnormal{s}}}
\def\rt{{\textnormal{t}}}
\def\ru{{\textnormal{u}}}
\def\rv{{\textnormal{v}}}
\def\rw{{\textnormal{w}}}
\def\rx{{\textnormal{x}}}
\def\ry{{\textnormal{y}}}
\def\rz{{\textnormal{z}}}

\def\rvepsilon{{\mathbf{\epsilon}}}
\def\rvtheta{{\mathbf{\theta}}}
\def\rva{{\mathbf{a}}}
\def\rvb{{\mathbf{b}}}
\def\rvc{{\mathbf{c}}}
\def\rvd{{\mathbf{d}}}
\def\rve{{\mathbf{e}}}
\def\rvf{{\mathbf{f}}}
\def\rvg{{\mathbf{g}}}
\def\rvh{{\mathbf{h}}}
\def\rvu{{\mathbf{i}}}
\def\rvj{{\mathbf{j}}}
\def\rvk{{\mathbf{k}}}
\def\rvl{{\mathbf{l}}}
\def\rvm{{\mathbf{m}}}
\def\rvn{{\mathbf{n}}}
\def\rvo{{\mathbf{o}}}
\def\rvp{{\mathbf{p}}}
\def\rvq{{\mathbf{q}}}
\def\rvr{{\mathbf{r}}}
\def\rvs{{\mathbf{s}}}
\def\rvt{{\mathbf{t}}}
\def\rvu{{\mathbf{u}}}
\def\rvv{{\mathbf{v}}}
\def\rvw{{\mathbf{w}}}
\def\rvx{{\mathbf{x}}}
\def\rvy{{\mathbf{y}}}
\def\rvz{{\mathbf{z}}}

\def\erva{{\textnormal{a}}}
\def\ervb{{\textnormal{b}}}
\def\ervc{{\textnormal{c}}}
\def\ervd{{\textnormal{d}}}
\def\erve{{\textnormal{e}}}
\def\ervf{{\textnormal{f}}}
\def\ervg{{\textnormal{g}}}
\def\ervh{{\textnormal{h}}}
\def\ervi{{\textnormal{i}}}
\def\ervj{{\textnormal{j}}}
\def\ervk{{\textnormal{k}}}
\def\ervl{{\textnormal{l}}}
\def\ervm{{\textnormal{m}}}
\def\ervn{{\textnormal{n}}}
\def\ervo{{\textnormal{o}}}
\def\ervp{{\textnormal{p}}}
\def\ervq{{\textnormal{q}}}
\def\ervr{{\textnormal{r}}}
\def\ervs{{\textnormal{s}}}
\def\ervt{{\textnormal{t}}}
\def\ervu{{\textnormal{u}}}
\def\ervv{{\textnormal{v}}}
\def\ervw{{\textnormal{w}}}
\def\ervx{{\textnormal{x}}}
\def\ervy{{\textnormal{y}}}
\def\ervz{{\textnormal{z}}}

\def\rmA{{\mathbf{A}}}
\def\rmB{{\mathbf{B}}}
\def\rmC{{\mathbf{C}}}
\def\rmD{{\mathbf{D}}}
\def\rmE{{\mathbf{E}}}
\def\rmF{{\mathbf{F}}}
\def\rmG{{\mathbf{G}}}
\def\rmH{{\mathbf{H}}}
\def\rmI{{\mathbf{I}}}
\def\rmJ{{\mathbf{J}}}
\def\rmK{{\mathbf{K}}}
\def\rmL{{\mathbf{L}}}
\def\rmM{{\mathbf{M}}}
\def\rmN{{\mathbf{N}}}
\def\rmO{{\mathbf{O}}}
\def\rmP{{\mathbf{P}}}
\def\rmQ{{\mathbf{Q}}}
\def\rmR{{\mathbf{R}}}
\def\rmS{{\mathbf{S}}}
\def\rmT{{\mathbf{T}}}
\def\rmU{{\mathbf{U}}}
\def\rmV{{\mathbf{V}}}
\def\rmW{{\mathbf{W}}}
\def\rmX{{\mathbf{X}}}
\def\rmY{{\mathbf{Y}}}
\def\rmZ{{\mathbf{Z}}}

\def\ermA{{\textnormal{A}}}
\def\ermB{{\textnormal{B}}}
\def\ermC{{\textnormal{C}}}
\def\ermD{{\textnormal{D}}}
\def\ermE{{\textnormal{E}}}
\def\ermF{{\textnormal{F}}}
\def\ermG{{\textnormal{G}}}
\def\ermH{{\textnormal{H}}}
\def\ermI{{\textnormal{I}}}
\def\ermJ{{\textnormal{J}}}
\def\ermK{{\textnormal{K}}}
\def\ermL{{\textnormal{L}}}
\def\ermM{{\textnormal{M}}}
\def\ermN{{\textnormal{N}}}
\def\ermO{{\textnormal{O}}}
\def\ermP{{\textnormal{P}}}
\def\ermQ{{\textnormal{Q}}}
\def\ermR{{\textnormal{R}}}
\def\ermS{{\textnormal{S}}}
\def\ermT{{\textnormal{T}}}
\def\ermU{{\textnormal{U}}}
\def\ermV{{\textnormal{V}}}
\def\ermW{{\textnormal{W}}}
\def\ermX{{\textnormal{X}}}
\def\ermY{{\textnormal{Y}}}
\def\ermZ{{\textnormal{Z}}}

\def\vzero{{\bm{0}}}
\def\vone{{\bm{1}}}
\def\vmu{{\bm{\mu}}}
\def\vtheta{{\bm{\theta}}}
\def\va{{\bm{a}}}
\def\vb{{\bm{b}}}
\def\vc{{\bm{c}}}
\def\vd{{\bm{d}}}
\def\ve{{\bm{e}}}
\def\vf{{\bm{f}}}
\def\vg{{\bm{g}}}
\def\vh{{\bm{h}}}
\def\vi{{\bm{i}}}
\def\vj{{\bm{j}}}
\def\vk{{\bm{k}}}
\def\vl{{\bm{l}}}
\def\vm{{\bm{m}}}
\def\vn{{\bm{n}}}
\def\vo{{\bm{o}}}
\def\vp{{\bm{p}}}
\def\vq{{\bm{q}}}
\def\vr{{\bm{r}}}
\def\vs{{\bm{s}}}
\def\vt{{\bm{t}}}
\def\vu{{\bm{u}}}
\def\vv{{\bm{v}}}
\def\vw{{\bm{w}}}
\def\vx{{\bm{x}}}
\def\vy{{\bm{y}}}
\def\vz{{\bm{z}}}

\def\evalpha{{\alpha}}
\def\evbeta{{\beta}}
\def\evepsilon{{\epsilon}}
\def\evlambda{{\lambda}}
\def\evomega{{\omega}}
\def\evmu{{\mu}}
\def\evpsi{{\psi}}
\def\evsigma{{\sigma}}
\def\evtheta{{\theta}}
\def\eva{{a}}
\def\evb{{b}}
\def\evc{{c}}
\def\evd{{d}}
\def\eve{{e}}
\def\evf{{f}}
\def\evg{{g}}
\def\evh{{h}}
\def\evi{{i}}
\def\evj{{j}}
\def\evk{{k}}
\def\evl{{l}}
\def\evm{{m}}
\def\evn{{n}}
\def\evo{{o}}
\def\evp{{p}}
\def\evq{{q}}
\def\evr{{r}}
\def\evs{{s}}
\def\evt{{t}}
\def\evu{{u}}
\def\evv{{v}}
\def\evw{{w}}
\def\evx{{x}}
\def\evy{{y}}
\def\evz{{z}}

\def\mA{{\bm{A}}}
\def\mB{{\bm{B}}}
\def\mC{{\bm{C}}}
\def\mD{{\bm{D}}}
\def\mE{{\bm{E}}}
\def\mF{{\bm{F}}}
\def\mG{{\bm{G}}}
\def\mH{{\bm{H}}}
\def\mI{{\bm{I}}}
\def\mJ{{\bm{J}}}
\def\mK{{\bm{K}}}
\def\mL{{\bm{L}}}
\def\mM{{\bm{M}}}
\def\mN{{\bm{N}}}
\def\mO{{\bm{O}}}
\def\mP{{\bm{P}}}
\def\mQ{{\bm{Q}}}
\def\mR{{\bm{R}}}
\def\mS{{\bm{S}}}
\def\mT{{\bm{T}}}
\def\mU{{\bm{U}}}
\def\mV{{\bm{V}}}
\def\mW{{\bm{W}}}
\def\mX{{\bm{X}}}
\def\mY{{\bm{Y}}}
\def\mZ{{\bm{Z}}}
\def\mBeta{{\bm{\beta}}}
\def\mPhi{{\bm{\Phi}}}
\def\mLambda{{\bm{\Lambda}}}
\def\mSigma{{\bm{\Sigma}}}

\newcommand{\tens}[1]{\bm{\mathsfit{#1}}}
\def\tA{{\tens{A}}}
\def\tB{{\tens{B}}}
\def\tC{{\tens{C}}}
\def\tD{{\tens{D}}}
\def\tE{{\tens{E}}}
\def\tF{{\tens{F}}}
\def\tG{{\tens{G}}}
\def\tH{{\tens{H}}}
\def\tI{{\tens{I}}}
\def\tJ{{\tens{J}}}
\def\tK{{\tens{K}}}
\def\tL{{\tens{L}}}
\def\tM{{\tens{M}}}
\def\tN{{\tens{N}}}
\def\tO{{\tens{O}}}
\def\tP{{\tens{P}}}
\def\tQ{{\tens{Q}}}
\def\tR{{\tens{R}}}
\def\tS{{\tens{S}}}
\def\tT{{\tens{T}}}
\def\tU{{\tens{U}}}
\def\tV{{\tens{V}}}
\def\tW{{\tens{W}}}
\def\tX{{\tens{X}}}
\def\tY{{\tens{Y}}}
\def\tZ{{\tens{Z}}}

\def\gA{{\mathcal{A}}}
\def\gB{{\mathcal{B}}}
\def\gC{{\mathcal{C}}}
\def\gD{{\mathcal{D}}}
\def\gE{{\mathcal{E}}}
\def\gF{{\mathcal{F}}}
\def\gG{{\mathcal{G}}}
\def\gH{{\mathcal{H}}}
\def\gI{{\mathcal{I}}}
\def\gJ{{\mathcal{J}}}
\def\gK{{\mathcal{K}}}
\def\gL{{\mathcal{L}}}
\def\gM{{\mathcal{M}}}
\def\gN{{\mathcal{N}}}
\def\gO{{\mathcal{O}}}
\def\gP{{\mathcal{P}}}
\def\gQ{{\mathcal{Q}}}
\def\gR{{\mathcal{R}}}
\def\gS{{\mathcal{S}}}
\def\gT{{\mathcal{T}}}
\def\gU{{\mathcal{U}}}
\def\gV{{\mathcal{V}}}
\def\gW{{\mathcal{W}}}
\def\gX{{\mathcal{X}}}
\def\gY{{\mathcal{Y}}}
\def\gZ{{\mathcal{Z}}}

\def\sA{{\mathbb{A}}}
\def\sB{{\mathbb{B}}}
\def\sC{{\mathbb{C}}}
\def\sD{{\mathbb{D}}}
\def\sF{{\mathbb{F}}}
\def\sG{{\mathbb{G}}}
\def\sH{{\mathbb{H}}}
\def\sI{{\mathbb{I}}}
\def\sJ{{\mathbb{J}}}
\def\sK{{\mathbb{K}}}
\def\sL{{\mathbb{L}}}
\def\sM{{\mathbb{M}}}
\def\sN{{\mathbb{N}}}
\def\sO{{\mathbb{O}}}
\def\sP{{\mathbb{P}}}
\def\sQ{{\mathbb{Q}}}
\def\sR{{\mathbb{R}}}
\def\sS{{\mathbb{S}}}
\def\sT{{\mathbb{T}}}
\def\sU{{\mathbb{U}}}
\def\sV{{\mathbb{V}}}
\def\sW{{\mathbb{W}}}
\def\sX{{\mathbb{X}}}
\def\sY{{\mathbb{Y}}}
\def\sZ{{\mathbb{Z}}}

\def\emLambda{{\Lambda}}
\def\emA{{A}}
\def\emB{{B}}
\def\emC{{C}}
\def\emD{{D}}
\def\emE{{E}}
\def\emF{{F}}
\def\emG{{G}}
\def\emH{{H}}
\def\emI{{I}}
\def\emJ{{J}}
\def\emK{{K}}
\def\emL{{L}}
\def\emM{{M}}
\def\emN{{N}}
\def\emO{{O}}
\def\emP{{P}}
\def\emQ{{Q}}
\def\emR{{R}}
\def\emS{{S}}
\def\emT{{T}}
\def\emU{{U}}
\def\emV{{V}}
\def\emW{{W}}
\def\emX{{X}}
\def\emY{{Y}}
\def\emZ{{Z}}
\def\emSigma{{\Sigma}}

\newcommand{\etens}[1]{\mathsfit{#1}}
\def\etLambda{{\etens{\Lambda}}}
\def\etA{{\etens{A}}}
\def\etB{{\etens{B}}}
\def\etC{{\etens{C}}}
\def\etD{{\etens{D}}}
\def\etE{{\etens{E}}}
\def\etF{{\etens{F}}}
\def\etG{{\etens{G}}}
\def\etH{{\etens{H}}}
\def\etI{{\etens{I}}}
\def\etJ{{\etens{J}}}
\def\etK{{\etens{K}}}
\def\etL{{\etens{L}}}
\def\etM{{\etens{M}}}
\def\etN{{\etens{N}}}
\def\etO{{\etens{O}}}
\def\etP{{\etens{P}}}
\def\etQ{{\etens{Q}}}
\def\etR{{\etens{R}}}
\def\etS{{\etens{S}}}
\def\etT{{\etens{T}}}
\def\etU{{\etens{U}}}
\def\etV{{\etens{V}}}
\def\etW{{\etens{W}}}
\def\etX{{\etens{X}}}
\def\etY{{\etens{Y}}}
\def\etZ{{\etens{Z}}}

\newcommand{\normlzero}{L^0}
\newcommand{\normlone}{L^1}
\newcommand{\normltwo}{L^2}
\newcommand{\normlp}{L^p}
\newcommand{\normmax}{L^\infty}

\def\sR{{\mathbb{R}}}

\newcommand\blfootnote[1]{%
  \begingroup
  \renewcommand\thefootnote{}\footnote{#1}%
  \addtocounter{footnote}{-1}%
  \endgroup
}
\blfootnote{*Equal contribution.}
\begin{abstract}
Photo-realistic visualization and animation of expressive human faces have been a long standing challenge. 
3D face modeling methods provide parametric control but generates unrealistic images, on the other hand, generative 2D models like GANs (Generative Adversarial Networks) output photo-realistic face images, but lack explicit control. 
Recent methods gain partial control, either by attempting to disentangle different factors in an unsupervised manner, or by adding control post hoc to a pre-trained model. 
Unconditional GANs, however, may entangle factors that are hard to undo later. 
We condition our generative model on pre-defined control parameters to encourage disentanglement in the generation process.
Specifically, we condition StyleGAN2 on FLAME, a generative 3D face model. 
While conditioning on FLAME parameters yields unsatisfactory results, we find that conditioning on rendered FLAME geometry and photometric details works well.
This gives us a generative 2D face model named GIF (Generative Interpretable Faces) that offers FLAME's parametric control. 
Here, interpretable refers to the semantic meaning of different parameters.
Given FLAME parameters for shape, pose, expressions, parameters for appearance, lighting, and an additional style vector, GIF outputs photo-realistic face images. 
We perform an AMT based perceptual study to quantitatively and qualitatively evaluate how well GIF follows its conditioning.
The code, data, and trained model are publicly available for research purposes at \url{http://gif.is.tue.mpg.de}.
\end{abstract}

\section{Introduction}

The ability to generate a person's face has several uses in computer graphics and computer vision that include constructing a personalized avatar for multimedia applications, face recognition and face analysis. To be widely useful, a generative model must offer control over the generative factors such as expression, pose, shape, lighting, skin tone, etc.
Early work focuses on learning a low dimensional representation of human faces using \ac{PCA} spaces~\cite{Cootes:ECCV:1998:Activeapp,Cootes:PAMI:2001:ActiveApp,Cootes:CVIU:1995:ActiveShape,Craw:BMVC:1991:Parameterising,Turk:JCN:1991:Eigenfaces} or higher-order tensor generalizations~\cite{Vasilescu:ECCV:2002:Multilinear}.
Although they provide some semantic control, these methods use linear transformations in the pixel-domain to model facial variation, resulting in blurry images. 
Further, effects of rotations are not well parameterized by linear transformations in 2D, resulting in poor image quality.

To overcome this, Blanz and Vetter~\cite{Blanz:CGIT:1999:Morphable} introduced a statistical 3D morphable model, of facial shape and appearance.
Such a statistical face model (e.g.~\cite{Cao:TVCG:2013:Facewarehouse,FLAME:SiggraphAsia2017,Paysan2009_BFM}) is used to manipulate shape, expression, or pose of the facial 3D geometry. This is then combined with a texture map (e.g.~\cite{Saito2017}) and rendered to an image. 
Rendering of a 3D face model lacks photo-realism due to the difficulty in modeling hair, eyes, and the mouth cavity (i.e., teeth or tongue), along with the absence of facial details like wrinkles in the geometry of the facial model. 
Further difficulties arise from subsurface scattering of facial material. These factors affect the photo-realism of the final rendering.

On the other hand, generative adversarial networks (GANs) have recently shown great success in generating photo-realistic face images at high resolution~\cite{Karras2018progressive}.
Methods like StyleGAN~\cite{karras2019style} or StyleGAN2~\cite{karras2019analyzing} even provide high-level control over factors like pose or identity when trained on face images. 
However, these controlling factors are often entangled, and they are unknown prior to training.
The control provided by these models does not allow the independent change of attributes like facial appearance, shape (e.g. length, width, roundness, etc.) or facial expression (e.g. raise eyebrows, open mouth, etc.). 
Although these methods have made significant progress in image quality, the provided control is not sufficient for graphics applications.

In short, we close the control-quality  gap. However, we find that a naive conditional version of StyleGAN2 yields  unsatisfactory results. We overcome this problem by rendering the geometric and photo-metric details of the FLAME mesh to an image and by using these as conditions instead. This design combination results in a generative 2D face model called GIF (Generative Interpretable Faces) that produces photo-realistic images with explicit control over face shape, head and jaw pose, expression, appearance, and illumination (Figure~\ref{fig:teaser_with_lighting}).

Finally we remark that the community lacks an automated metric to effectively evaluate continuous conditional generative models. To this end we derive a quantitative score from a comparison-based perceptual study using an anonymous group of participants to facilitate future model comparisons.

In summary, our main contributions are 
1) a generative 2D face model with FLAME~\cite{FLAME:SiggraphAsia2017} control,
2) use of FLAME renderings as conditioning for better association, 
3) use of a texture consistency constraint to improve disentanglement,
4) providing a quantitative comparison mechanism.

\section{Related Work}

\qheading{Generative 3D face models: }
Representing and manipulating human faces in 3D have a long standing history dating back almost five decades to the parametric 3D face model of Parke~\cite{Parke:1974:Parametric}.
Blanz and Vetter~\cite{Blanz:CGIT:1999:Morphable} propose a 3D morphable model (3DMM), the first generative 3D face model that uses linear subspaces to model shape and appearance variations. 
This has given rise to a variety of 3D face models to model facial shape~\cite{Booth2018_LSFM, Dai2017, Paysan2009_BFM}, shape and expression~\cite{Amberg2008,Blanz2003,BolkartWuhrer2015,Cao:TVCG:2013:Facewarehouse,COMA:ECCV18,Vlasic2005}, shape, expression and head pose~\cite{FLAME:SiggraphAsia2017}, localized facial details~\cite{Brunton2014, neog2016interactive} and wrinkle details~\cite{Golovinskiy2006,Shin2014}.
However, renderings of these models do not reach photo-realism due to the lack of high-quality textures.

To overcome this, Saito et al.~\cite{Saito2017} introduce high-quality texture maps, and Slossberg et al.~\cite{Slossberg2018} and Gecer et al.~\cite{Gecer:CVPR:2019:Ganfit} train GANs to synthesize textures with high-frequency details.
While these works enhance the realism when being rendered by covering more texture details, they only model the face region (i.e. ignore hair, teeth, tongue, eyelids, eyes, etc.) required for photo-realistic rendering. 
While separate part-based generative models of hair~\cite{Hu2017_hair,Saito2018_hair,Wei2018_hair}, eyes~\cite{Berard2014_eyes}, eyelids~\cite{Bermano2015_eyelids}, ears~\cite{Dai2018_ear}, teeth~\cite{Wu2016_teeth}, or tongue~\cite{Hewer2018} exist, combining these into a complete realistic 3D face model remains an open problem.

Instead of explicitly modeling all face parts, Gecer et al.~\cite{Gecer:ECCV:2018:Semi} use image-to-image translation to enhance the realism of images rendered from a 3D face mesh. 
Nagano et al.~\cite{Nagano:ACM:2018:paGAN} generate dynamic textures that allow synthesizing expression dependent mouth interior and varying eye gaze.
Despite significant progress of generative 3D face models~\cite{Brunton:CVIU:2014:Review,Egger:arxiv:2019:3dMroph}, they still lack photo-realism.

Our approach, in contrast, combines the semantic control of generative 3D models with the image synthesis ability of generative 2D models. 
This allows us to generate photo-realistic face images, including hair, eyes, teeth, etc. with explicit 3D face model controls.

\qheading{Generative 2D face models: }
Early parametric 2D face models like Eigenfaces~\cite{SirovichKirby1987,Turk:JCN:1991:Eigenfaces}, Fisherfaces~\cite{Belhumeur1997eigenfaces}, Active Shape Models~\cite{Cootes:CVIU:1995:ActiveShape}, or Active Appearance Models~\cite{Cootes:ECCV:1998:Activeapp} parametrize facial shape and appearance in images with linear spaces.
Tensor faces~\cite{Vasilescu:ECCV:2002:Multilinear} generalize these linear models to higher-order, generating face images with multiple independent factors like identity, pose, or expression.
Although these models provided some semantic control, they produced blurry and unrealistic images.
 
StyleGAN~\cite{karras2019style}, a member of broad category of GAN~\cite{Goodfellow2014_GAN} models, extends Progressive-GAN~\cite{Karras2018progressive} by incorporating a style vector to gain partial control over the image generation, which is broadly missing in such models.
However, the semantics of these controls are interpreted only post-training. Hence, it is possible that desired controls might not be present at all. 
InterFaceGAN~\cite{Shen:arxiv:2019:Interpreting} and StyleRig~\cite{tewari2020stylerig} aim to gain control over a pre-trained StyleGAN~\cite{karras2019style}.
InterFaceGAN~\cite{Shen:arxiv:2019:Interpreting} identifies hyper-planes that separate positive and negative semantic attributes in a GAN's latent space.
However, this requires categorical attributes for every kind of control, making it not suitable for a variety of aspects e.g. facial shape, expression etc. 
Further, many attributes might not be linearly separable. 
StyleRig~\cite{tewari2020stylerig} learns mappings between 3DMM parameters and the parameter vectors of each StyleGAN layer. 
StyleRig learns to edit StyleGAN parameters and thereby controls the generated image, with 3DMM parameters.  The setup is mainly tailored towards face editing or face reenactment tasks. 
GIF, in contrast, provides full generative control over the image generation process (similar to regular GANs) but with semantically meaningful control over shape, pose, expression, appearance, lighting, and style. 

CONFIG~\cite{kowalski2020config} leverages synthetic images to get ground truth control parameters and leverages real images to make the image generation look more realistic. However, generated images still lack photo-realism. 

HoloGAN~\cite{nguyen2019hologan} randomly applies rigid transformations to learnt features during training, which provide explicit control over 3D rotations in the trained model.
While this is feasible for global transformations, it remains unclear how to extend this to transformations of local parts or parameters like facial shape or expression. 
Similarly, IterGAN~\cite{Galama:CVIU:2019:Itergans} also only models rigid rotation of generated objects using a GAN, but these rotations are restricted to 2D transformations. 

Further works use variational autoencoders~\cite{razavi2019generating,van2017neural} and flow-based methods~\cite{kingma2018glow} to generate images. These provide controllability, but do not reach the image quality of GANs.

\qheading{Facial animation: }  
A large body of work focuses on face editing or facial animation, which can be grouped into 3D model-based approaches (e.g.~\cite{Geng2019,Kim:ACM:2018:Deep,Lombardi2018,Thies:ACM:2019:Deferred,Thies2016_Face2Face,Ververas2019}) or 3D model-free methods (e.g.~\cite{Bansal:ECCV:2018:Recycle,Pumarola:ECCV:2018:Ganimation,tripathy2020icface,Wu:ECCV:2018:Reenactgan,Zakharov:ICCV:2019:Few})

Thies et al.~\cite{Thies2016_Face2Face} build a subject specific 3DMM from a video, reenact this model with expression parameters from another sequence, and blend the rendered mesh with the target video. 
Follow-up work use similar 3DMM-based retargeting techniques but replace the traditional rendering pipeline, or parts of it, with learnable components~\cite{Kim:ACM:2018:Deep,Thies:ACM:2019:Deferred}.
Ververas and Zafeiriou~\cite{Ververas2019} (Slider-GAN) and Geng et al.~\cite{Geng2019} propose image-to-image translation models that, given an image of a particular subject, condition the face editing process on 3DMM parameters.
Lombardi et al.~\cite{Lombardi2018} learn a subject-specific autoencoder of facial shape and appearance from high-quality multi-view images that allow animation and photo-realistic rendering. 
Like GIF, all these methods use explicit control of a pre-trained 3DMM or learn a 3D face model to manipulate or animate faces in images, but in contrast to GIF, they are unable to generate new identities.

Zakharov et al.~\cite{Zakharov:ICCV:2019:Few} use image-to-image translation to animate a face image from 2D landmarks, ReenactGAN~\cite{Wu:ECCV:2018:Reenactgan} and Recycle-GAN~\cite{Bansal:ECCV:2018:Recycle} transfer facial movement from a monocular video to a target person. 
Pumarola et al. \cite{Pumarola:ECCV:2018:Ganimation} learn a GAN conditioned on facial action units for control over facial expressions. 
None of these methods provide explicit control over a 3D face representation. 

All methods discussed above are task specific, i.e., they are dedicated towards manipulating or animating faces, while GIF, in contrast, is a generative 2D model that is able to generate new identities, and also provides control of a 3D face model.
Further, most of the methods are trained on video data~\cite{Bansal:ECCV:2018:Recycle,Kim:ACM:2018:Deep,Lombardi2018,Thies:ACM:2019:Deferred,Thies2016_Face2Face,Zakharov:ICCV:2019:Few}, in contrast to GIF which is trained from static images only.
Regardless, GIF can be used to generate facial animations. 
\section{Preliminaries}
\label{prilims}
\qheading{GANs and conditional GANs:} 
GANs are a class of neural networks where a generator $\generator$ and a discriminator $\discriminator$ have opposing objectives. Namely, the discriminator estimates the probability of its input to be a generated sample, as opposed to a natural random sample from the training set, while the generator tries to make this task as hard as possible. 
This is extended in the case of conditional GANs~\cite{mirza2014conditional}. The objective function in such a setting is given as
\begin{equation}
    \begin{split}
        \min_G\max_D V(\discriminator, \generator) = \expt_{x\sim p(x)}[\log\discriminator(x|c)] + \\ \expt_{z\sim p(z)}[\log(1-\discriminator(\generator(z|c))],
    \end{split}
    \label{eq:c_gan_obj}
\end{equation}
where $c$ is the conditioning variable.
Although sound in an ideal setting, this formulation suffers a major drawback. Specifically under incomplete data regime, independent conditions tend to influence each other. In Section \ref{sec:cond_representation}, we discuss this phenomenon in detail.

\subsection{StyleGAN2}
\label{sec:stylegan}
StyleGAN2~\cite{Karras2019stylegan2}, a revised version of StyleGAN ~\cite{Karras2018progressive}, produces photo-realistic face images at $1024 \times 1024$ resolution.
Similar to StyleGAN, StyleGAN2, is controlled by a style vector $\rvz$. 
This vector is first transformed by a mapping network of $8$ fully connected layers to $\rvw$, which then transforms the activations of the progressively growing resolution blocks using adaptive instance normalization (AdaIN) layers.
Although StyleGAN2 provides some high-level control, it still lacks explicit and semantically meaningful control. 
Our work addresses this shortcoming by distilling a conditional generative model out of StyleGAN2 and combining this with inputs from FLAME.

\subsection{FLAME}
FLAME is a publicly available 3D head model~\cite{FLAME:SiggraphAsia2017}, $M(\shapecoeff, \posecoeff, \expcoeff): \mathbb{R}^{\shapedim \times \posedim \times \expdim}\rightarrow \mathbb{R}^{\numverts \times 3}$, which given parameters for facial shape $\shapecoeff \in \mathbb{R}^{300}$, pose $\posecoeff \in \mathbb{R}^{15}$ (i.e. axis-angle rotations for global rotation and rotations around joints for neck, jaw, and eyeballs), and facial expression $\expcoeff \in \mathbb{R}^{100}$ outputs a mesh with $\numverts = 5023$ vertices. 
We further transfer the appearance space of Basel Face Model~\cite{Paysan2009_BFM} parametrized by $\albedocoeffs \in \mathbb{R}^{\albedodim}$ into the FLAME's UV layout to augment it with a texture space.
We use the same subset of FLAME parameters as RingNet~\cite{Sanyal:CVPR:2019:RingNet}, namely $6$ pose coefficients for global rotation and jaw rotation, $100$ shape, and $50$ expression parameters, and we use $50$ parameters for appearance.
We use rendered FLAME meshes as the conditioning signal in GIF.

\section{Method}
\label{sec:method}
\begin{figure*}
    \centering
    \includegraphics[trim=0 10 0 10, clip, width=.9\linewidth]{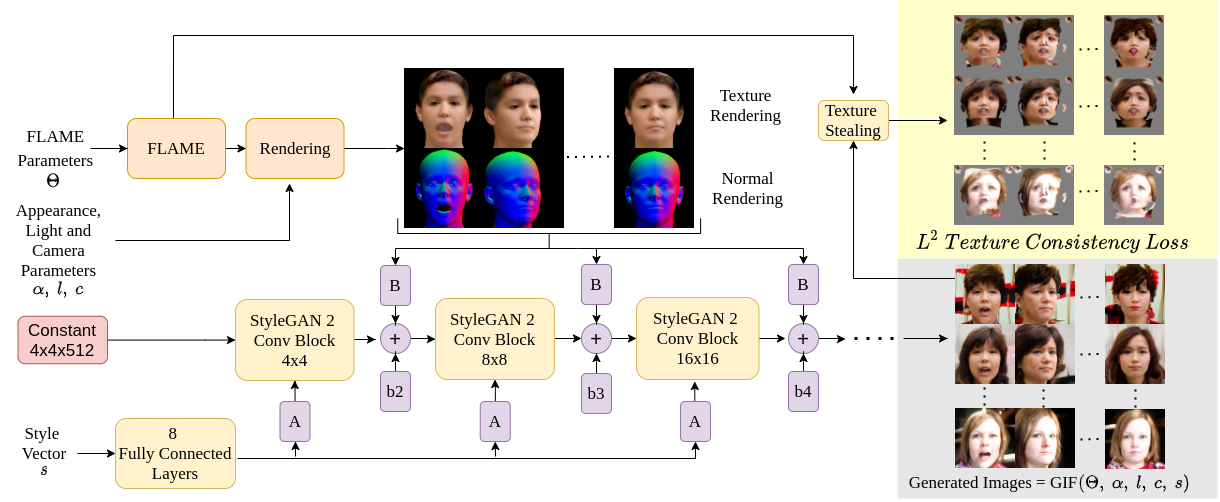}
    \caption{Our generator architecture is based  on StyleGAN2~\cite{Karras2019stylegan2}, $A$ is a learned affine transform, and $B$ stands for per-channel scaling. We make several key changes, such as introducing 3D model generated condition through the noise injection channels and introduce texture consistency loss. We refer to the process of projecting the generated image onto the FLAME mesh to obtain an incomplete texture map as \textit{texture stealing}.}
    \label{figure:gif_model}
\end{figure*}

\qheading{Goal:}
Our goal is to learn a generative 2D face model controlled by a parametric 3D face model.
Specifically, we seek a mapping GIF$(\Theta,\albedocoeffs,\lighting,\camera,\style): \sR^{156+50+27+3+512} \rightarrow \sR^{\imgsize \times \imgsize \times 3}$, that given FLAME parameters $\Theta = \{\shapecoeff,\posecoeff,\expcoeff\} \in \sR^{156}$, parameters for appearance $\albedocoeffs \in \sR^{50}$, spherical harmonics lighting $\lighting \in \sR^{27}$, camera $\camera \in \sR^{3}$ (2D translation and isotropic scale of a weak-perspective camera), and style $\style \in \sR^{512}$, generates an image of resolution $\imgsize\times\imgsize$.

Here, the FLAME parameters control all aspects related to the geometry, appearance and lighting parameters control the face color (i.e. skin tone, lighting, etc.), while the style vector $\style$ controls all factors that are not described by the FLAME geometry and appearance parameters, but are required to generate photo-realistic face images (e.g. hairstyle, background, etc.).

\subsection{Training data}
\label{sub_sec:dat_pre_process}

Our data set consists of Flickr images (FFHQ) introduced in StyleGAN~\cite{karras2019style} and their corresponding FLAME parameters, appearance parameters, lighting parameters, and camera parameters.
We obtain these using DECA~\cite{DECA2020}, a publicly available monocular 3D face regressor. 
In total, we use about 65,500 FFHQ images, paired with the corresponding parameters. 
The obtained FFHQ training parameters are available for research purposes. 

\subsection{Condition representation}
\label{sec:cond_representation}

\qheading{Condition cross-talk:} 
The vanilla conditional GAN formulation as described in Section~\ref{prilims} does not encode any semantic factorization of the conditional probability distributions that might exist in the nature of the problem. 
Consider a situation where the true data depends upon two independent generating factors $c_1$ and $c_2$, i.e. the true generation process of our data is  $x \sim P(x | c_1, c_2)$ where $P(x, c_1, c_2) = P(x | c_1, c_2) \cdot P(c_1) \cdot P(c_2)$. 
Ideally, given complete data (often infinite) and a perfect modeling paradigm, this factorization should emerge automatically. However, in practice, neither of these can be assumed. 
Hence, the representation of the condition highly influences the way it gets associated with the output. 
We refer to this phenomenon of independent conditions influencing each other as -- \emph{condition cross-talk}. 
Inductive bias introduced by condition representation in the context of conditional cross-talk is empirically evaluated in Section~\ref{sec:experiments}.

\qheading{Pixel-aligned conditioning:} 
Learning the rules of graphical projection (orthographic or perspective) and the notion of occlusion as part of the generator is wasteful if explicit 3D geometry information is present, as it approximates classical rendering operations, which can be done learning-free and are already part of several software packages (e.g.~\cite{Loper:ECCV:2014,Ravi2020pytorch3d}).
Hence, we provide the generator with explicit knowledge of the 3D geometry by conditioning it with renderings from a classical renderer. 
This makes pixel-localized association between the FLAME conditioning signal and the generated image possible. 
We find that, although a vanilla conditional GAN achieves comparable image quality, GIF learns to better obey the given condition. %

We condition GIF on two renderings, one provides pixel-wise color information (referred to as texture rendering), and the other provides information about geometry (referred to as normal rendering). 
Normal renderings are obtained by rendering the mesh with a color-coded map of the surface normals $\mathbf{n} = \mathbf{N}(M(\shapecoeff, \posecoeff, \expcoeff))$. 

Both renderings use a scaled orthographic projection with the camera parameters provided with each training image.
For the color rendering, we use the provided inferred lighting and appearance parameter.
The texture and the normal renderings are concatenated along the color channel and used to condition the generator as shown in Figure~\ref{figure:gif_model}. 
As demonstrated in Section~\ref{sec:experiments}, this conditioning mechanism helps reduce condition cross-talk.

\subsection{\ourmodel architecture}
\label{sec:model_architecture}

The model architecture of \ourmodel is based on StyleGAN2~\cite{Karras2019stylegan2}, with several key modifications, discussed as follows. An overview is shown in Figure~\ref{figure:gif_model}.

\qheading{Style embedding:} 
Rendering the textured FLAME mesh does not consider hair, mouth cavity (i.e., teeth or tongue), and fine-scale details like wrinkles and pores. 
Generating a realistic face image, however, requires these factors to be considered. Hence we introduce a style vector $\style \in \sR^{512}$ to model these factors.
Note that original StyleGAN2 has a similar vector $\rvz$ with the same dimensionality, however, instead of drawing random samples from a standard normal $\mathcal{N}(0, I)$ distribution (as common in GANs), we assign a random but unique vector for every image. 
This is motivated by the key observation that in the FFHQ dataset~\cite{karras2019style}, each identity mostly occurs only once and mostly has a unique background. 
Hence, if we use a dedicated vector for each image using, e.g., an embedding layer, we will encode an inductive bias for this vector to capture background and appearance specific information.

\qheading{Noise channel conditioning:} 
StyleGAN and StyleGAN2 insert random noise images at each resolution level into the generator, which mainly contributes to local texture changes. 
We replace this random noise by the concatenated textured and normal renderings from the FLAME model, and insert scaled versions of these renderings at different resolutions into the generator.
This is motivated by the observation that varying FLAME parameters, and therefore varying FLAME renderings, should have direct, pixel-aligned influence on the generated images. 

\subsection{Texture consistency}
\label{Sec:texture_consistency}
Since GIF's generation is based on an underlying 3D model, we can further constrain the generator by introducing a texture consistency loss optimized during training. 
We first generate a set of new FLAME parameters by randomly interpolating between the parameters within a mini batch. Next, we generate the corresponding images with the same style embedding $\style$, appearance $\albedocoeffs $ and lighting parameters $\lighting$ by an additional forward pass through the model. Finally, the  corresponding FLAME meshes are projected onto the  generated images to get a partial texture map (also referred to as `texture stealing'). 
To enforce pixel-wise consistency, we apply an $\normltwo$ loss on the difference between pairs of texture maps, considering only pixels for which the corresponding 3D point on the mesh surface is visible in both the generated images. 
We find that this texture consistency loss improves the parameter association (see Section~\ref{sec:experiments}).
\section{Experiments}
\label{sec:experiments}

\subsection{Qualitative evaluation}

\begin{figure}
\begin{center}
	\begin{tabular}{c c c}
	        \begin{turn}{90}\thead{GIF($\shapecoeff_1$,$\posecoeff_1$, \\ $\expcoeff_1$,$\albedocoeffs_1$,$\lighting_1$,  \\ $\camera_1$,$\style_1$)}\end{turn} & 
            \includegraphics[height=0.065\textheight]{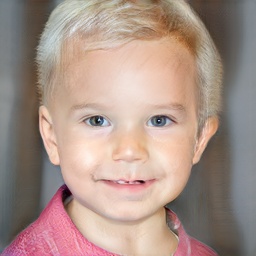}
            \includegraphics[height=0.065\textheight]{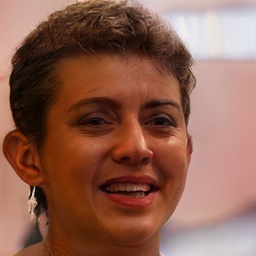}
            \includegraphics[height=0.065\textheight]{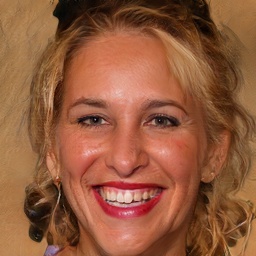}
            \includegraphics[height=0.065\textheight]{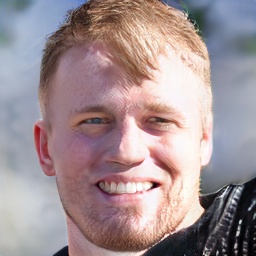}\\
        
	        \begin{turn}{90}\thead{GIF($\shapecoeff_1$,$\posecoeff_1$, \\ $\expcoeff_1$,$\albedocoeffs_1$,$\lighting_1$,  \\ $\camera_1$, $\color{red}\style_2$)}\end{turn} & 
            \includegraphics[height=0.065\textheight]{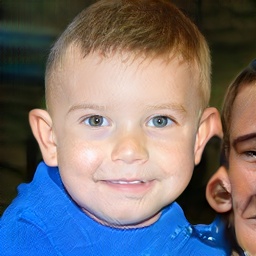}
            \includegraphics[height=0.065\textheight]{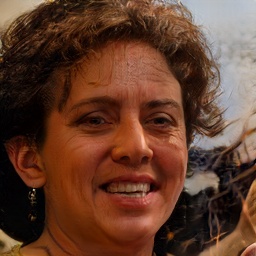}
            \includegraphics[height=0.065\textheight]{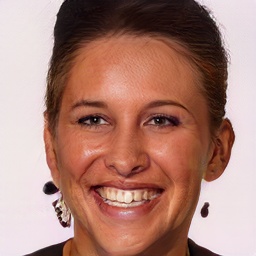}
            \includegraphics[height=0.065\textheight]{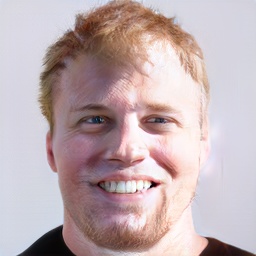}\\
 
	        \begin{turn}{90}\thead{$GIF$($\color{red}\shapecoeff_2$,$\posecoeff_1$, \\ $\expcoeff_1,\albedocoeffs_1,\lighting_1$,  \\ $\camera_1$,$\color{red}\style_2$)}\end{turn} & 
            \includegraphics[height=0.065\textheight]{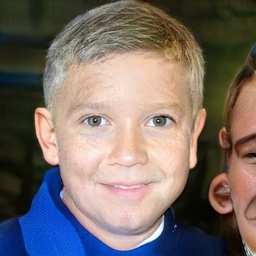}
            \includegraphics[height=0.065\textheight]{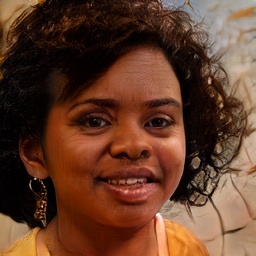}
            \includegraphics[height=0.065\textheight]{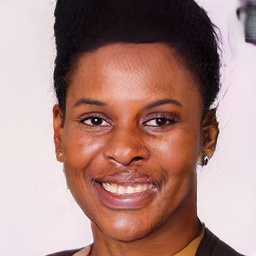}
            \includegraphics[height=0.065\textheight]{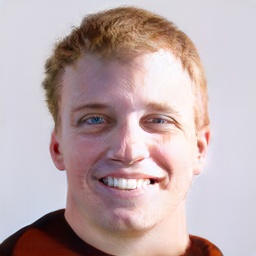}\\

	        \begin{turn}{90}\thead{GIF($\color{red}\shapecoeff_2$,$\posecoeff_1$, \\ $\color{red}\expcoeff_2$,$\albedocoeffs_1,\lighting_1$,  \\ $\camera_1$,$\color{red}\style_2$)}\end{turn} & 
            \includegraphics[height=0.065\textheight]{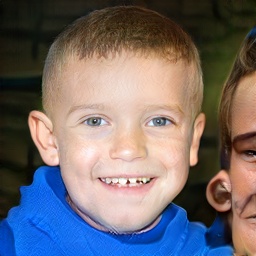}
            \includegraphics[height=0.065\textheight]{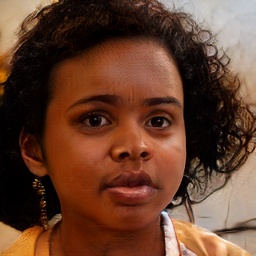}
            \includegraphics[height=0.065\textheight]{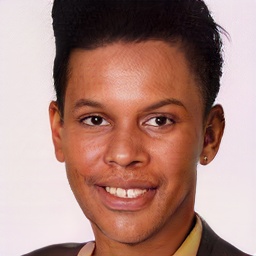}
            \includegraphics[height=0.065\textheight]{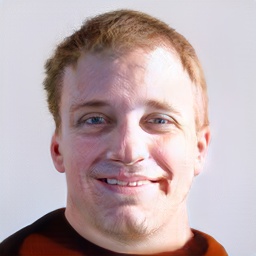}\\

	        \begin{turn}{90}\thead{GIF($\color{red}\shapecoeff_2,\posecoeff_2$, \\ $\color{red}\expcoeff_2$,$\albedocoeffs_1,\lighting_1$,  \\ $\color{red}\camera_2,\style_2$)}\end{turn} & 
            \includegraphics[height=0.065\textheight]{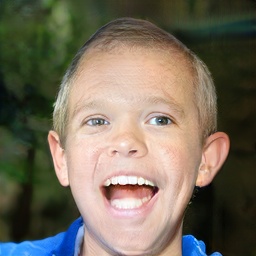}
            \includegraphics[height=0.065\textheight]{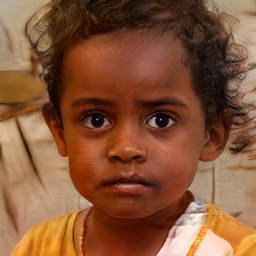}
            \includegraphics[height=0.065\textheight]{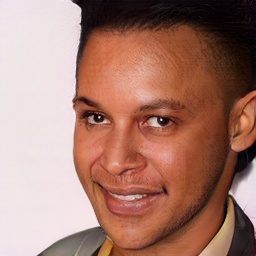}
            \includegraphics[height=0.065\textheight]{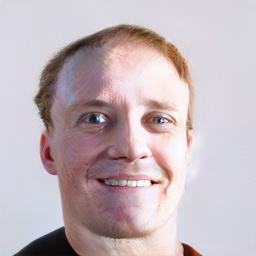}\\
            
	        \begin{turn}{90}\thead{GIF($\color{red}\shapecoeff_2,\posecoeff_2$, \\ $\color{red}\expcoeff_2,\albedocoeffs_2$,$\lighting_1$,  \\ $\color{red}\camera_2,\style_2$)}\end{turn} & 
            \includegraphics[height=0.065\textheight]{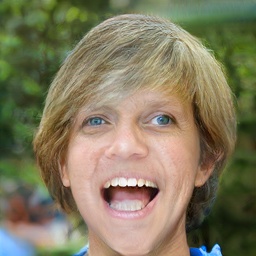}
            \includegraphics[height=0.065\textheight]{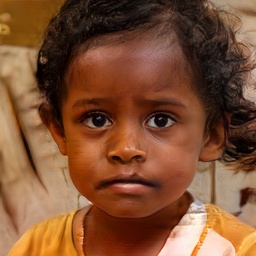}
            \includegraphics[height=0.065\textheight]{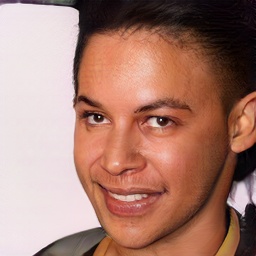}
            \includegraphics[height=0.065\textheight]{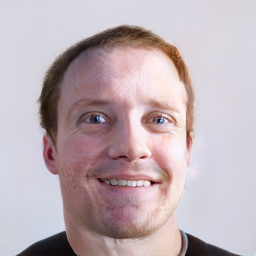}\\
            
	        \begin{turn}{90}\thead{GIF($\color{red}\shapecoeff_2,\posecoeff_2$, \\ $\color{red}\expcoeff_2,\albedocoeffs_2,\lighting_2$,  \\ $\color{red}\camera_2,\style_2$)}\end{turn} & 
            \includegraphics[height=0.065\textheight]{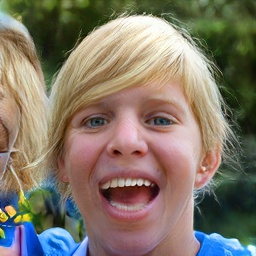}
            \includegraphics[height=0.065\textheight]{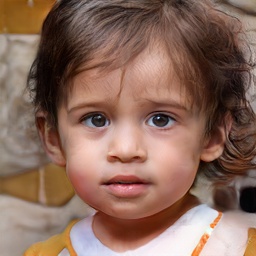}
            \includegraphics[height=0.065\textheight]{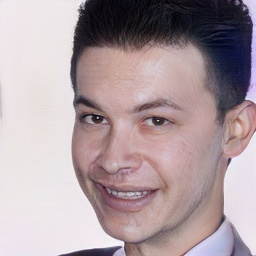}
            \includegraphics[height=0.065\textheight]{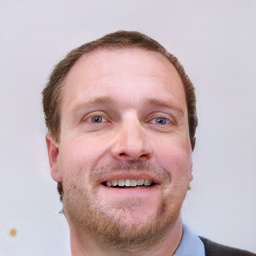}\\
    \end{tabular}
    \caption{Impact of individual parameters, when being exchanged between two different generated images one at a time. From top to bottom we progressively exchange style, shape, expression, head and jaw pose, appearance, and lighting of the two parameter sets. We progressively change the color of the parameter symbol that is effected in every row to red.}
    \label{fig:cond_swap}
\end{center}
\vspace{-3mm}
\end{figure}

\qheading{Condition influence:}
As described in Section~\ref{sec:method}, GIF is parametrized by FLAME parameters $\Theta = \{\shapecoeff, \posecoeff, \expcoeff\}$, appearance parameters $\albedocoeffs$, lighting parameters $\lighting$, camera parameters $\camera$ and a style vector $\style$. 
Figure~\ref{fig:cond_swap} shows the influence of each individual set of parameters by progressively exchanging one type of parameter in each row. 
The top and bottom rows show GIF generated images for two sets of parameters, randomly chosen from the training data. 

Exchanging style (row 2) most noticeably changes hair style, clothing color, and the background.
Shape (row 3) is strongly associated to the person's identity among other factors. 
The expression parameters (row 4) control the facial expression, best visible around the mouth and cheeks. 
The change in pose parameters (row 5) affects the orientation of the head (i.e. head pose) and the extent of the mouth opening (jaw pose).
Finally, appearance (row 6) and lighting (row 7) change the skin color and the lighting specularity.

\begin{figure*}
    \centerline{
       \includegraphics[height=0.122\textheight]{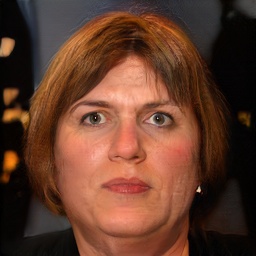}
       \includegraphics[height=0.122\textheight]{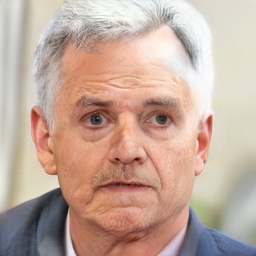}
       \includegraphics[height=0.122\textheight]{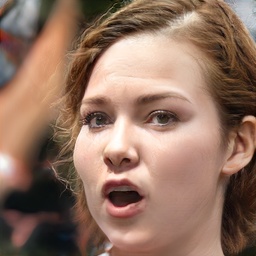}
       \includegraphics[height=0.122\textheight]{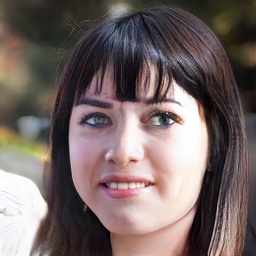}
       \includegraphics[height=0.122\textheight]{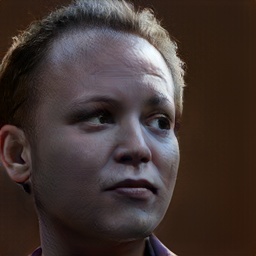}
       \includegraphics[height=0.122\textheight]{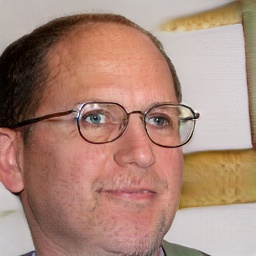}
    }
    \centerline{
      \includegraphics[height=0.122\textheight]{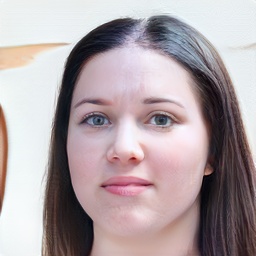}
      \includegraphics[height=0.122\textheight]{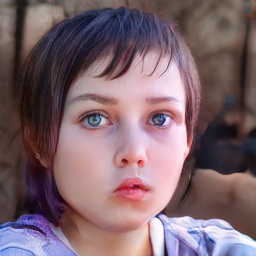}
      \includegraphics[height=0.122\textheight]{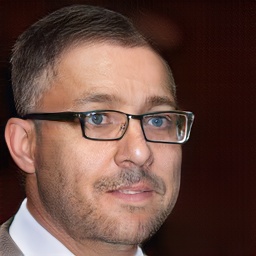}
      \includegraphics[height=0.122\textheight]{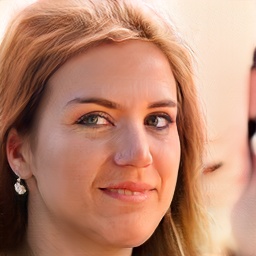}
      \includegraphics[height=0.122\textheight]{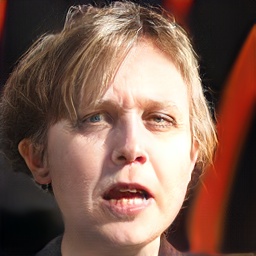}
      \includegraphics[height=0.122\textheight]{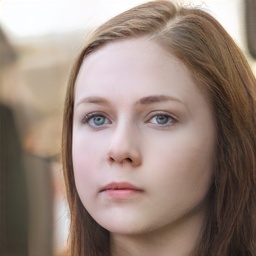}
    }        
    \caption{Images obtained by randomly sampling FLAME, appearance parameters, style parameters, and lighting parameters. Specifically, for shape, expression, and appearance parameters, we sample parameters of the first three principal components from a standard normal distribution and keep all other parameters at zero. 
    We sample pose from a uniform distribution in $[-\pi/8,+\pi/8]$ (head pose) for rotation around the y-axis, and $[0,+\pi/12]$ (jaw pose) around the x-axis}
    \label{fig:random_samples}
\end{figure*}

\qheading{Random sampling:}
To further evaluate GIF qualitatively, we sample FLAME parameters, appearance parameters, lighting parameters, and style embeddings and generate random images, shown in Figure~\ref{fig:random_samples}. 
For shape, expression, and appearance parameters, we sample parameters of the first three principal components from a standard normal distribution and keep all other parameters at zero. 
For pose, we sample from a uniform distribution in $[-\pi/8,+\pi/8]$ (head pose) for rotation around the y-axis, and $[0,+\pi/12]$ (jaw pose) around the x-axis.  
For lighting parameters and style embeddings, we choose random samples from the set of training parameters.
Figure~\ref{fig:random_samples} shows that GIF produces photo-realistic images of different identities with a large variation in shape, pose, expression, skin color, and age. 
Figure~\ref{fig:teaser_with_lighting} further shows rendered FLAME meshes for generated images, demonstrating that GIF generated images are well associated with the FLAME parameters.

\qheading{Speech driven animation:}
As GIF uses FLAME's parametric control, it can directly be combined with existing FLAME-based application methods such as VOCA~\cite{VOCA2019}, which animates a face template in FLAME mesh topology from speech.
For this, we run VOCA for a speech sequence, fit FLAME to the resulting meshes, and use these parameters to drive GIF for different appearance embeddings (see Figure~\ref{fig:voca_seq}).
For more qualitative results and the full animation sequence, see the supplementary video. 

\begin{figure}
    \centerline{
        \includegraphics[height=0.067\textheight]{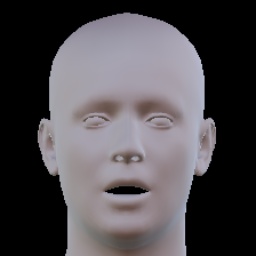}
        \includegraphics[height=0.067\textheight]{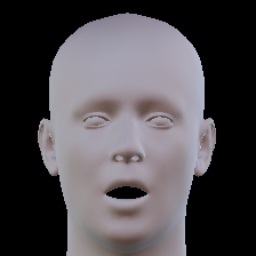}
        \includegraphics[height=0.067\textheight]{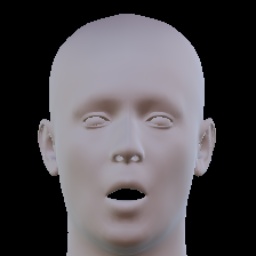}
        \includegraphics[height=0.067\textheight]{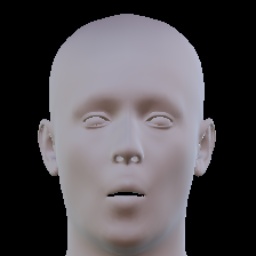}
        \includegraphics[height=0.067\textheight]{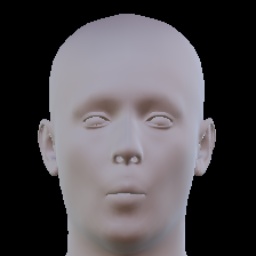}
    }
    \centerline{
        \includegraphics[height=0.067\textheight]{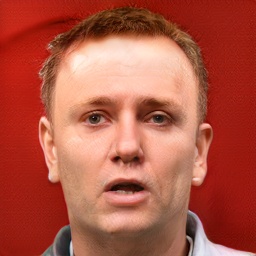}
        \includegraphics[height=0.067\textheight]{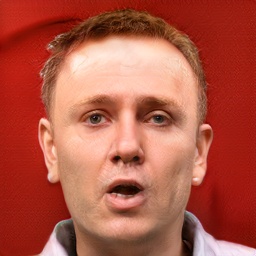}        
        \includegraphics[height=0.067\textheight]{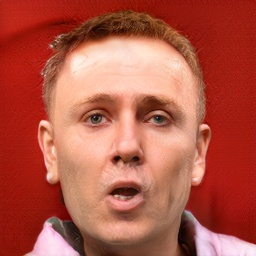}
        \includegraphics[height=0.067\textheight]{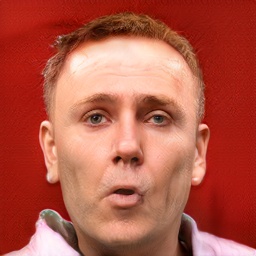}
        \includegraphics[height=0.067\textheight]{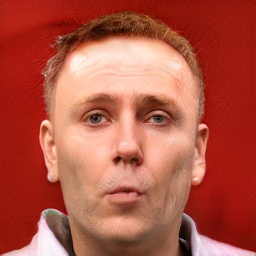}
    }
    \centerline{
        \includegraphics[height=0.067\textheight]{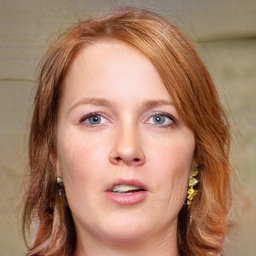}
        \includegraphics[height=0.067\textheight]{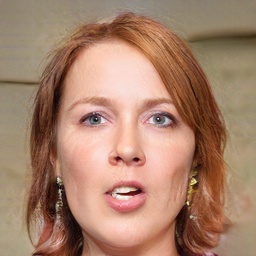}        
        \includegraphics[height=0.067\textheight]{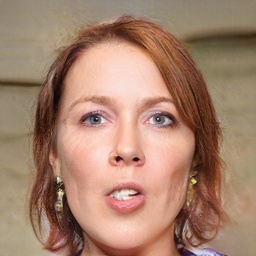}
        \includegraphics[height=0.067\textheight]{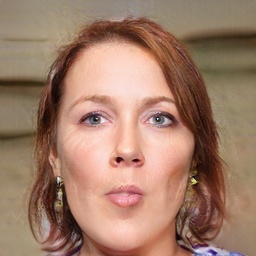}
        \includegraphics[height=0.067\textheight]{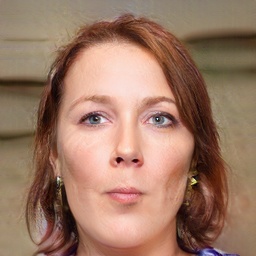}
    }
    \caption{Combination of GIF and an existing speech-driven facial animation method by generating face images for FLAME parameters obtained from VOCA~\cite{VOCA2019}. Sample frames to highlight jaw pose variation. Please see the supplementary video for the full animation.}
    \vspace{-2mm}
    \label{fig:voca_seq}
\end{figure}

\subsection{Quantitative evaluation}
\label{sec:quantitative_eval}

We conduct two \ac{AMT} studies to quantitatively evaluate i) the effects of ablating individual model parts, and ii) the disentanglement of geometry and style. 
We compare GIF in total with $4$ different ablated versions, namely \textit{vector conditioning}, \textit{no texture interpolation}, \textit{normal rendering}, and \textit{texture rendering conditioning}.
For the \textit{vector conditioning} model, we directly provide the FLAME parameters as a $236$ dimensional real valued vector. 
In the \textit{no texture interpolation} model, we drop the texture consistency during training.
The \textit{normal rendering} and \textit{texture rendering conditioning} models condition the generator and discriminator only on the normal rendering or texture rendering, respectively, while removing the other rendering.
The supplementary video shows examples for both user studies.

\qheading{Ablation experiment:}
Participants see three images, a reference image in the center, which shows a rendered FLAME mesh, and two generated images to the left and right in random order. 
Both images are generated from the same set of parameters, one using GIF, another an ablated GIF model.
Participants then select the generated image that corresponds best with the reference image.
Table~\ref{table:modelA-vs_mdl_B-Ais_chosen_more_often} shows that with the texture consistency loss, normal rendering, and texture rendering conditioning, GIF performs slightly better than without each of them. 
Participants tend to select GIF generated results over a vanilla conditional StyleGAN2 (please refer to our supplementary material for details on the architecture). 
Furthermore, Figure~\ref{table:fid_over_varying_sigma} quantitatively evaluates the image quality with FID scores, indicating that all models produce similar high-quality results.

\begin{table}
    \centering
    \begin{tabular}{c c c c c}
        \toprule
         & \thead{Vector \\ cond.} & \thead{No Texture \\ interpolation} & \thead{Normal rend.\\conditioning} & \thead{Texture rend.\\ conditioning} \\
        \midrule
         GIF & 89.4\% & 51.1\% & 55.8\% & 51.7\% \\
         \bottomrule
    \end{tabular}
    \caption{AMT ablation experiment. Preference percentage of GIF generated images over ablated models and vector conditioning model. Participants were instructed to pay particular attention to shape, pose, and expression and ignore image quality.}
    \label{table:modelA-vs_mdl_B-Ais_chosen_more_often}
\end{table}

\begin{table}
    \centering
    \begin{tabular}{c c c c c c}
        \toprule
        GIF & \thead{Vector \\ cond.} & \thead{No Texture \\ interpolation} & \thead{Normal rend. \\conditioning} & \thead{Texture rend.\\ conditioning} \\
        \midrule
         \textbf{8.94}  & 10.34 & 11.71 & 9.89 & 11.28 \\
         \bottomrule
    \end{tabular}
    \caption{FID scores of images generated by GIF and ablated models (lower is better). Note that this score only evaluates image quality and does not judge how well the models obey the underlying FLAME conditions.}
    \label{table:fid_over_varying_sigma}
\end{table}

\begin{figure}[t]
    \includegraphics[width=\linewidth]{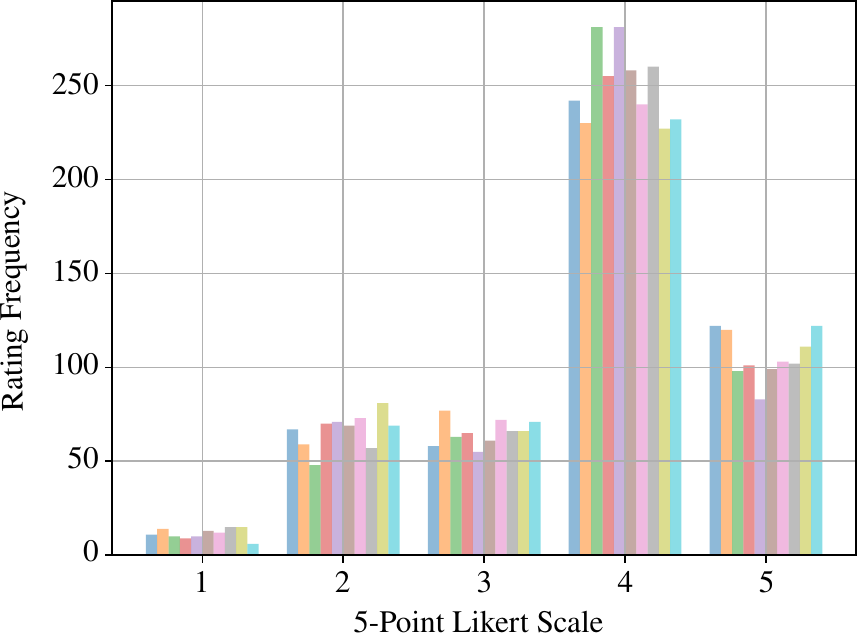}
    \caption{Preference frequency of styles on a 5-Point Likert Scale.  Note that almost all style vectors, represented with different colors here get a similar distribution of likeness ratings indicating that they do not influence the perceived FLAME conditioning.}
    \label{fig:flam_asso_likert}
\end{figure}

\qheading{Geometry-style disentanglement:} 
In this experiment, we study the entanglement between the style vector and FLAME parameters. 
We find qualitatively that the style vector mostly controls aspects of the image that are not influenced by FLAME parameters like background, hairstyle, etc. (see Figure~\ref{fig:cond_swap}). 
To evaluate this quantitatively, we randomly pair style vectors and FLAME parameters and conduct a perceptual study with AMT.
Participants see a rendered FLAME image and GIF generated images with the same FLAME parameters but with a variety of different style vectors. 
Participants then rate the similarity of the generated image to shape, pose and expression of the FLAME rendering on a standard 5-Point Likert scale (i.e. 1: Strongly Disagree, 2. Disagree, 3. Neither agree nor disagree, 4. Agree, 5. Strongly Agree). 
We use $10$ randomized styles $\style$ and $500$ random FLAME parameters totaling to $5000$ images. 
We find that the majority of the participants agree that generated images and FLAME rendering are similar, irrespective of the style (see Figure~\ref{fig:flam_asso_likert}).

\qheading{Re-inference error:} 
In the spirit of DiscoFaceGAN~\cite{deng2020disentangled}, we run DECA~\cite{DECA2020} on the generated images and compute the Root Mean Square Error (RMSE) between the FLAME face vertices of the input parameters and the inferred parameters as reported in Table~\ref{table:mse_table}. 
We generate a population of $1000$ images by randomly sampling one of the shape, pose and expression latent spaces while holding the rest of the generating factors to their neutral. Thus we find an association error for individual factors.
\begin{table}[h!]
    \centering
    \begin{tabular}{c c c c}
        \toprule
        Model & Shape & Expression & Pose \\
        \midrule
         Vector cond. & 3.43 mm & 23.05 mm & 29.69 mm\\
         GIF & \textbf{3.02} mm & \textbf{5.00} mm & \textbf{5.61} mm\\
         \bottomrule
    \end{tabular}
    \caption{To evaluate shape error, we generate 1024 random faces from GIF and re-infer their shape using DECA~\cite{DECA2020}. Using FLAME, we compute face region vertices twice: once with GIF's input condition and once with re-inferred parameters. Finally, the RMSE between these face region vertices are computed. The process is repeated for expression and pose.}
    \label{table:mse_table}
\end{table}

\section{Discussion}

GIF is trained on the FFHQ data-set and hence inherits some of its limitations, e.g. the images are roughly eye-centered. 
Although StyleGAN2 ~\cite{karras2019analyzing} has to some extent addressed this issue (among others), but failed to do so completely. 
Hence, rotations of the head look like an eye-centered rotation, which involves a combination of 3D rotation and translation as opposed to a pure neck centered rotation. 
Adapting GIF to use an architecture other than StyleGAN2  or training it on a different data set to improve image quality is subject to future work.

As FLAME renderings for conditioning GIF must be similarly eye-centered as the FFHQ training data. 
We compute a suitable camera parameter from given FLAME parameters so that the eyes are located roughly at the centre of the image plane with a fixed distance between the eyes. 
However, for profile view poses, this can not be met without an extreme zoomed in view. 
This often causes severe artifacts. 
Please see the supplementary material for examples. 

GIF requires a statistical 3D model, and a way to associate its parameters to a large data set of high-quality images. 
While `objects' like human bodies~\cite{pavlakos2019expressive} or animals~\cite{Zuffi2019} potentially fulfill these requirements, it remains unclear how to apply GIF to general object categories. 

Faulty image to 3D model associations stemming from the parameter inference method potentially degrade GIF's generation quality. 
One example is the ambiguity of lighting and appearance, which causes most color variations in the training data to be described by lighting variation rather than by appearance variation. GIF inherits these errors.

Finally, as GIF is solely trained from static images without multiple images per subject, generating images with varying FLAME parameters is not temporally consistent. 
As several unconditioned parts are only loosely correlated or uncorrelated to the condition (e.g. hair, mouth cavity, background, etc.), this results in jittery video sequences.
Training or refining GIF on temporal data with additional temporal constraints during training remains subject to future work.

\section{Conclusion}

We present GIF, a generative 2D face model, with high realism and with explicit control from FLAME, a statistical 3D face model. 
Given a data set of approximately 65,500 high-quality face images with associated FLAME model parameters for shape, global pose, jaw pose, expression, and appearance parameters, GIF learns to generate realistic face images that associate with them. 
Our key insight is that conditioning a generator network on explicit information rendered from a 3D face model allows us to decouple shape, pose, and expression variations within the trained model. 
Given a set of FLAME parameters associated with an image, we render the corresponding FLAME mesh twice, once with color-coded normal details, once with an inferred texture, and insert these as condition to the generator. 
We further add a loss that enforces consistency in texture for the reconstruction of different FLAME parameters for the same appearance embedding. 
This encourages the network during training to disentangle appearance and FLAME parameters, and provides us with better temporal consistency when generating frames of FLAME sequences. Finally we devise a comparison-based perceptual study to evaluate continuous conditional generative models quantitatively.
\section{Acknowledgements}
We thank H. Feng for prepraring the training data, Y. Feng and S. Sanyal for support with the rendering and projection pipeline, and C. Köhler, A. Chandrasekaran, M. Keller, M. Landry, C. Huang, A. Osman and D. Tzionas for fruitful discussions, advice and proofreading. 
The work was partially supported by the International Max Planck Research School for Intelligent Systems (IMPRS-IS).

\section{Disclosure}
MJB has received research gift funds from Intel, Nvidia, Adobe, Facebook, and Amazon. 
While MJB is a part-time employee of Amazon, his research was performed solely at, and funded solely by, MPI. 
MJB has financial interests in Amazon and Meshcapade GmbH.
PG has received funding from Amazon Web Services for using their Machine Learning Services and GPU Instances. AR's research was performed solely at, and funded solely by, MPI.

\clearpage
\appendix
\newcommand\doubleplus{+\kern-1.3ex+\kern0.8ex}
\twocolumn[{
    \begin{center}
        \includegraphics[width=\linewidth]{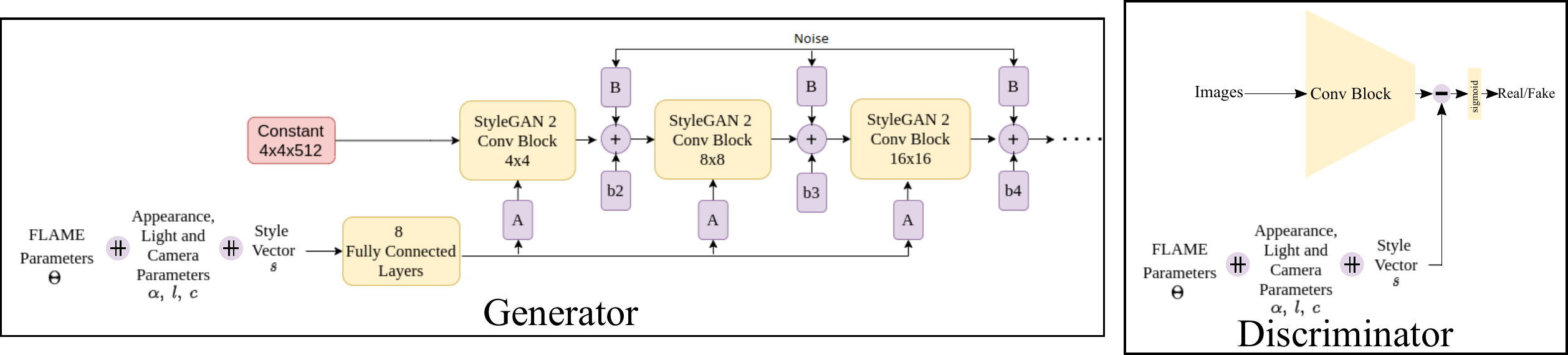}
        \captionof{figure}{Architecture of Vector conditioning model. Here $\doubleplus$ represents a concatenation.}
        \label{fig:vec_cond_architecture}
    \end{center}
}]
\section{Vector conditioning architecture}
Here we describe the model architecture of the vector condition model, used as one of the baseline models in Section 5.2 of the main paper. 
For this model, we pass the vector values conditioning parameters FLAME ($\shapecoeff, \posecoeff, \expcoeff, $), appearance ($\albedocoeffs$) and lighting ($\lighting$) as a $236$ dimensional vector through the dimensions of style vector of the original StyleGAN2 architecture as shown in Figure~\ref{fig:vec_cond_architecture}.
We further input the same conditioning vector to the discriminator at the last fully connected layer of the discriminator by subtracting it from the last layer activation. 

\begin{figure}[h]
    \centerline{
        \includegraphics[height=0.067\textheight]{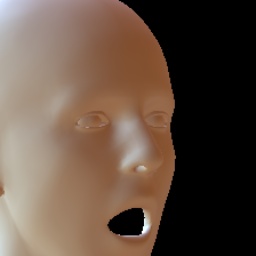}
        \includegraphics[height=0.067\textheight]{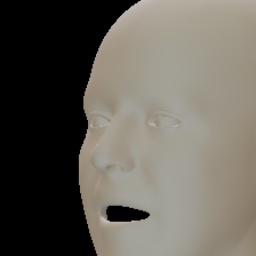}
        \includegraphics[height=0.067\textheight]{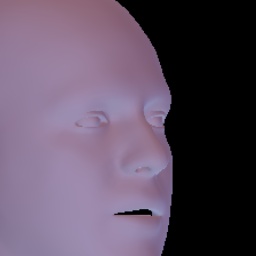}
        \includegraphics[height=0.067\textheight]{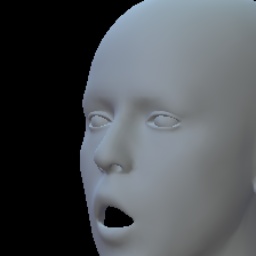}
        \includegraphics[height=0.067\textheight]{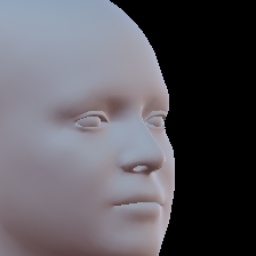}
    }
    \centerline{
        \includegraphics[height=0.067\textheight]{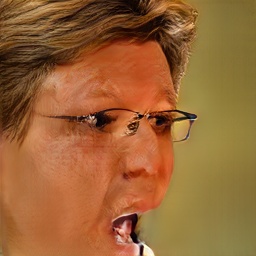}
        \includegraphics[height=0.067\textheight]{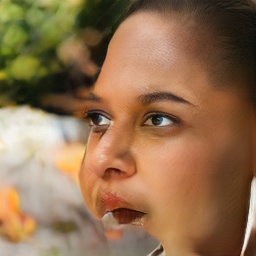}
        \includegraphics[height=0.067\textheight]{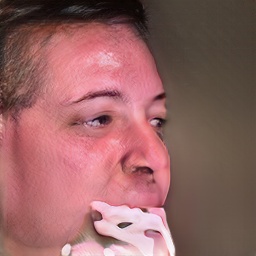}
        \includegraphics[height=0.067\textheight]{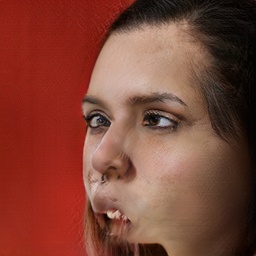}
        \includegraphics[height=0.067\textheight]{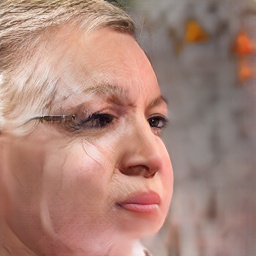}
    }
    \centerline{
        \includegraphics[height=0.067\textheight]{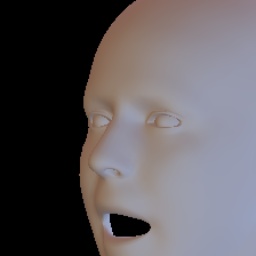}
        \includegraphics[height=0.067\textheight]{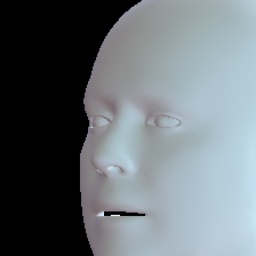}
        \includegraphics[height=0.067\textheight]{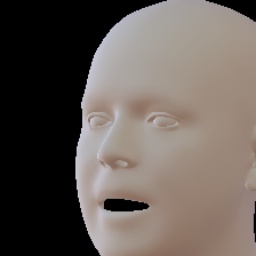}
        \includegraphics[height=0.067\textheight]{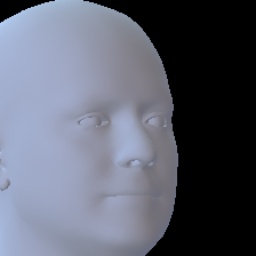}
        \includegraphics[height=0.067\textheight]{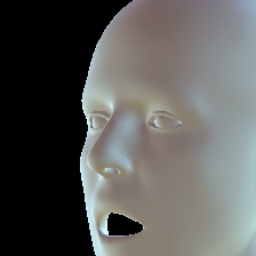}
    }
    \centerline{
        \includegraphics[height=0.067\textheight]{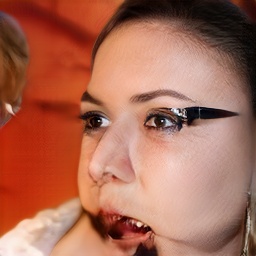}
        \includegraphics[height=0.067\textheight]{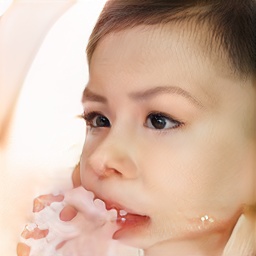}
        \includegraphics[height=0.067\textheight]{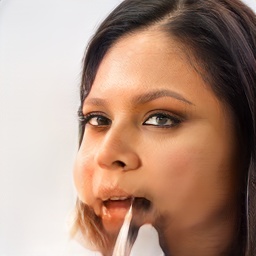}
        \includegraphics[height=0.067\textheight]{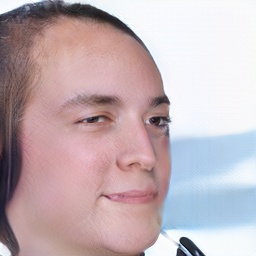}
        \includegraphics[height=0.067\textheight]{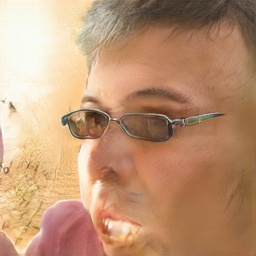}
    }    
    \captionof{figure}{In order for the eyes to be places at a given pixel location a profile view causes the face image to be highly zoomed in. This causes the generated images to become unrealistic.}
    \label{fig:too_zoomed_in_view}
\end{figure}

\section{Image centering for extreme rotations}

As discussed in Section 6 of the main paper, GIF produces artifacts for extreme head poses close to profile view. 
This is due to the pixel alignment of the FLAME renderings and the generated images, which requires the images to be similarly eye-centered as the FFHQ training data. 
For profile views however it is unclear how the centering within the training data was achieved. 
The centering strategy used in GIF causes a zoom in for profile views, effectively cropping parts of the face, and hence the generator struggles to generate realistic images as shown in Figure~\ref{fig:too_zoomed_in_view}.

\end{document}